\documentclass[journal]{IEEEtran}

\usepackage{ifpdf}

\usepackage{cite}

\ifCLASSINFOpdf
\usepackage[pdftex]{graphicx}
\graphicspath{{../pdf/}{../jpeg/}}
\DeclareGraphicsExtensions{.pdf,.jpeg,.png}
\else
\usepackage[dvips]{graphicx}
\graphicspath{{../eps/}}
\DeclareGraphicsExtensions{.eps}
\fi

\usepackage{amsmath}
\usepackage{algorithmic}
\usepackage{colortbl}
\usepackage{xcolor}
\usepackage{array}
\usepackage{makecell,multirow,diagbox}
\ifCLASSOPTIONcompsoc
\usepackage[caption=false,font=normalsize,labelfont=sf,textfont=sf]{subfig}
\else
\usepackage[caption=false,font=footnotesize]{subfig}
\fi

\usepackage{fixltx2e}
\usepackage{stfloats}

\ifCLASSOPTIONcaptionsoff
\usepackage[nomarkers]{endfloat}
\let\MYoriglatexcaption\caption
\renewcommand{\caption}[2][\relax]{\MYoriglatexcaption[#2]{#2}}
\fi

\usepackage{bm}
\usepackage[colorlinks, citecolor=blue, linkcolor=blue]{hyperref}
\usepackage{threeparttable}
\usepackage{amsfonts}
\usepackage{algorithm}
\usepackage{textcomp}
\usepackage{url}
\usepackage{verbatim}
\usepackage{graphicx}
\usepackage{booktabs}
\usepackage{longtable}
\usepackage{multicol}
\usepackage{pifont}
\usepackage{wasysym}
\usepackage[square, comma, sort, numbers]{natbib}
\usepackage{bm}
\usepackage{amsmath}
\usepackage{amssymb}
\makeatletter

\newcommand{\Rmnum}[1]{\expandafter\@slowromancap\romannumeral #1@}

\newcommand{\valueUP}[2]{{#1}\fontsize{3.5pt}{\baselineskip}\selectfont{}}

\newcommand{\valueDown}[2]{{#1}\fontsize{3.5pt}{\baselineskip}\selectfont{}}
\newcommand{\valueRange}[2]{{#1}\fontsize{3.5pt}{\baselineskip}\selectfont{$\pm${{#2}}}}
\newcommand{\up}{$\textcolor{green}{\uparrow}$}
\newcommand{\down}{$\textcolor{red}{\downarrow}$}

\newcommand{\modifyRS}[1]{\textcolor{black}{#1}}
\newcommand{\modifyRF}[1]{\textcolor{black}{#1}}
\makeatother

\newcommand{\orcid}[1]{\href{https://orcid.org/#1}{\includegraphics[width=10pt]{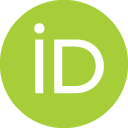}}}
\begin{document}
\title{Enhancing Cross-Dataset Performance of Distracted Driving Detection With Score Softmax Classifier And Dynamic Gaussian Smoothing Supervision}
%
\author{Cong~Duan {\orcid{0000-0001-8101-5447}},
	Zixuan~Liu \orcid{0009-0009-0335-5468},
	Jiahao~Xia \orcid{0000-0001-9628-9563}, \textit{Graduate Student Member, IEEE},\\
	Minghai~Zhang \orcid{0009-0004-0564-1728},
	Jiacai~Liao \orcid{0009-0000-9064-3322},
	Libo~Cao \orcid{0009-0009-6902-0994},

	\thanks{This work was
		supported by the National Natural Science Foundation of China under
		Grant 51621004. (\emph{Corresponding author: Libo Cao})}
	\thanks{C. Duan, Z. Liu, M. Zhang, L. Cao are all with State Key Laboratory of Advanced Design and Manufacturing for Vehicle Body, College of Mechanical and Vehicle Engineering, Hunan University, Changsha 410082, P.R. China (E-mail: duancong@hnu.edu.cn, lzx999@hnu.edu.cn, zmhai@hnu.edu.cn, hdclb@hnu.edu.cn) }
	\thanks{Jiahao Xia is with the Faculty of Engineering and IT, University of Technology Sydney, Ultimo, NSW 2007, Australia (e-mail: Jiahao.Xia@student.uts.edu.au).}
	\thanks{Jiacai Liao is with College of Automotive and Mechanical Engineering, ChangSha University of Science Technology, Changsha 410114, P. R. China (e-mail: ljc\_csust@csust.edu.cn).}
}

\markboth{Journal of \LaTeX\ Class Files,~Vol.~14, No.~8, August~2021}%
{Shell \MakeLowercase{\textit{et al.}}: A Sample Article Using IEEEtran.cls for IEEE Journals}


\maketitle
\begin{abstract}
Deep neural networks enable real-time monitoring of in-vehicle drivers, facilitating the timely prediction of distractions, fatigue, and potential hazards. This technology is now integral to intelligent transportation systems.
Recent research has exposed unreliable cross-dataset driver behavior recognition due to a limited number of data samples and background noise.
In this paper, we propose a Score-Softmax classifier, which reduces the model overconfidence by enhancing category independence.
Imitating the human scoring process, we designed a two-dimensional dynamic supervisory matrix consisting of one-dimensional Gaussian-smoothed labels. 
The dynamic loss descent direction and Gaussian smoothing increase the uncertainty of training to prevent the model from falling into noise traps.
Furthermore, we introduce a simple and convenient multi-channel information fusion method; it addresses the fusion issue among arbitrary Score-Softmax classification heads.
We conducted cross-dataset experiments using the SFDDD, AUCDD, and the 100-Driver datasets, demonstrating that Score-Softmax improves cross-dataset performance without modifying the model architecture. 
The experiments indicate that the Score-Softmax classifier reduces the interference of background noise, enhancing the robustness of the model.
It increases the cross-dataset accuracy by 21.34\%, 11.89\%, and 18.77\% on the three datasets, respectively.
\modifyRS{The code is publicly available at https://github.com/congduan-HNU/SSoftmax.}
\end{abstract}


\IEEEpeerreviewmaketitle

\section{INTRODUCTION}\label{section:1}
\IEEEPARstart{D}{istracted} driving represents a substantial contributor to road traffic risks, accounting for approximately 80\% of road collisions \cite{dingus2006100}. 
According to the U.S. National Highway Traffic Safety Administration (NHTSA), distracted driving encompasses ``any activity that diverts attention from driving, including manual, visual, and cognitive distractions \cite{NHTSA.org}."
These distracted actions result in substantial casualties and economic repercussions \cite{dong2010driver,wollmer2011online}. 
Annually, over a million fatalities and approximately 50 million injuries are reported in traffic accidents \cite{Report.org}.

\begin{figure}[!t]
	\centering
	\includegraphics[scale=0.5]{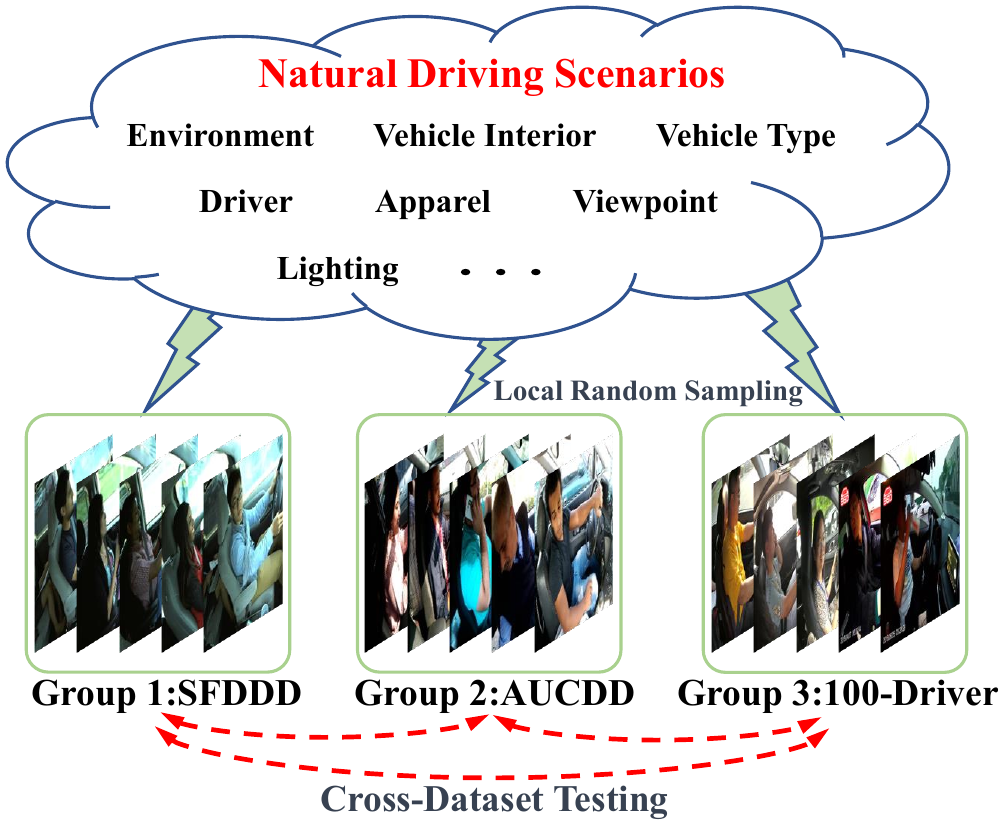}
	\caption{Similar noise features in in-vehicle camera samples can lead to local traps in the solution space. Prominent sources of noise, such as the windows, rearview mirrors, and the vehicle control panel, show significant variations across datasets. These differences are the primary causes for the local noise traps shown in Fig. \ref{Figure_optimize}.}
	\label{Figure_crossdataset}
\end{figure}
Driver monitoring systems (DMS) recognize driver distractions based on in-vehicle perception allowing for timely alerts or interventions to prevent traffic accidents caused by distraction \cite{kotserubaAttentionVisionBasedAssistive2022}.
For example, DMS can assess gaze based on head keypoints \cite{Lu_JHPFANet2023} and determine fatigue based on eye keypoints \cite{Li_Driver2021}.
Both are used to detect visual distractions and cannot deal with the issue of  manual distractions that involve driver behaviour.
The most widespread approach is to use end-to-end convolutional neural networks (CNNs) to extract the driver features and predict the probability of different behaviours \cite{Baheti_Computationally2020}.
Generally, the driver behaviours be defined as ``Drive Safely", ``Calling", ``Sending Text" and ``Drinking", etc. \cite{Abouelnaga_Realtime2018}.
Thus, detecting manual distractions based on driver behavior becomes a classification problem.

Due to the outstanding performance of CNNs in currently available distracted driving datasets \cite{StateFarm_Distracted,Abouelnaga_Realtime2018,Baheti_Detection2018,Eraqi_Driver2019}, the mainstream trend has shifted towards designing real-time and efficient model architectures \cite{Baheti_Computationally2020,Qin_Distracted2022,Li_Driver2022,liLearningAccurateSpeedy2022,Liu_Extremely2023,Mittal_CATCapsNet2023,Duan_FRNet2023}.
However, researchers have overlooked a critical issue that may hinder the technology from advancing towards large-scale practical applications. This issue is that testing methods fail to reflect the reliability in natural driving scenarios (NDS).
As depicted in Fig.\ref{Figure_crossdataset}, the sensor viewpoint, in-vehicle environment, and driver characteristics and attire vary randomly in NDS.
Since a dataset released by a single organization can be regarded as a small sample of the natural environment, testing methods applied solely on that dataset are insufficient to reflect the true performance of algorithms.

The practical issue of acquiring a reliable and extensive dataset of in-vehicle driver behavior poses considerable challenges. 
Not only is the large-scale deployment of data collection devices and manual annotation expensive, but drivers are also unwilling to compromise their privacy.
Therefore, a feasible way to measure the robustness of models is cross-dataset testing which was first conducted by Behera \emph{et al.} \cite{beheraLatentBodyPoseGuided2018}. 
Concretely, they found a CNN optimized on State Farm Distracted Driver Detection Dataset (SFDDD) \cite{StateFarm_Distracted} almost inoperative on American University in Cairo Distracted Driver (AUCDD) \cite{Abouelnaga_Realtime2018}, and vice versa.
In cross-dataset testing, the training set represents a small sample of NDS, while the test set reflects random events in NDS.
This suggests that large-scale driver distracted recognition based on CNN in NDS is highly unreliable.
Wang \emph{et al.} reported similar results regarding cross-view, cross-vehicle, and cross-modal scenes \cite{Wang_100Driver2023}.
Frank \emph{et al.} explained that this is because CNNs capture background noise instead of key features related to distracted driving \cite{zandamelaCrossdatasetPerformanceEvaluation2022}.

The One-Hot label leads to CNNs being overconfident and mistakenly treating noise as key features.
Label smoothing (LS) has been proven to prevent the model from becoming overly confident \cite{szegedyRethinkingInceptionArchitecture2016,mullerWhenDoesLabel2020}.
And it is also employed in distracted detection tasks \cite{alkanatDensityGuidedLabelSmoothing2022,wangFPTFineGrainedDetection2022,liLearningAccurateSpeedy2022,lyuDriverDistractedBehavior2023}.
Alternatively, adopting entirely different supervision approaches, such as triplet loss \cite{okonDetectingDistractedDriving2017}, contrastive loss \cite{huDriverAnomalyQuantification2023,yangQuantitativeIdentificationDriver2023}, and even unsupervised learning \cite{royUnsupervisedSparseNonnegative2022} is also viable.
Additionally, enhancing the raw data \cite{ouEnhancingDriverDistraction2020} or latent features \cite{Peng_TransDARC2022} is also a viable approaches.
In this paper, we further optimize the classification labels and supervision approach.
Specifically, we designed a two-dimensional classifier, called Score-Softmax, which untangles constraints between different categories and transforms the maximum probability prediction into a probability-weighted score prediction.
Moreover, we propose the dynamic Gaussian smoothing supervision (DGSS) method based on dynamic 2-D edge Gaussian distributed matrices, inspired by human rating patterns. 
This leads to oscillatory descent of loss, reducing the likelihood of falling into noise traps.
Additionally, we recommend a multi-channel data fusion strategy based on Gaussian distribution fusion, which is simpler and more convenient.
On the SFDDD, AUCDD, and 100-Driver \cite{Wang_100Driver2023} datasets, our strategy exhibited the superior cross-dataset performance with accuracy improvements of 21.34\%, 11.89\%, and 18.77\%, respectively. 
Our contributions can be summarized as follows:
\begin{itemize} 
	\item We designed the S-Softmax classifier to untangle constraints between different categories and transform the maximum probability prediction into a probability-weighted score prediction.
	\item To avoid falling into the background noise trap, we proposed DGSS to enhance intra-category label diversity and mitigate the constraints imposed on the model during training.
	\item Building upon S-Softmax, we propose a simpler, more convenient, and stable method for multi-channel information fusion.
\end{itemize}

The paper is organized as follows:
Section \ref{section:2} provides a literature review covering distracted driving detection, transfer learning, label smoothing and information fusion.
Section \ref{section:3} details the S-Softmax classifier, DGSS, and Gaussian fusion-based (GF) multi-channel feature fusion.
Section \ref{section:4} elaborates on the conducted experiments, while Section \ref{section:5} presents the experimental results.
Finally, Section \ref{section:6} concludes the research.

\section{RELATED WORKS}\label{section:2}
\subsection{Distracted Driving Detection}
Physiological sensors, including electroencephalogram (EEG) \cite{liTemporalSpatialDeepLearning2021}, electrocardiogram (ECG), and others, have been considered for monitoring driver distraction. However, invasive sensors may pose greater safety risks. Recently, computer vision breakthroughs have garnered significant attention for camera-based distracted driving detection solutions \cite{wangSurveyDriverBehavior2022}.

Among them, methods based on end-to-end CNNs are favored for their excellent performance and simplicity of implementation \cite{Baheti_Detection2018}. Moreover, models optimized through techniques like depthwise separable convolution or neural architecture search (NAS) exhibit higher real-time performance \cite{Baheti_Computationally2020,Duy-LinhNguyen_Driver2022,Li_Driver2022,Qin_Distracted2022,Liu_Extremely2023,Duan_FRNet2023}. Additionally, many researchers are exploring methods to enhance distracted driving recognition, such as key region detection \cite{Bera_Attend2021,wangDataAugmentationApproach2021} using cascaded CNNs or human body skeleton key point recognition \cite{liNovelSpatialTemporalGraph2019,luPoseawareDynamicWeighting2022,tanBidirectionalPostureAppearanceInteraction2022,liEffectiveMultiScaleFramework2023}.
Recent work has also investigated the performance of architectures based on the Multi-head Self-Attention mechanism (MSA) in distracted driving detection tasks \cite{wangSparseSpatiotemporalTransformer2023,zhangNovelDriverDistraction2023,maViTDDMultiTaskVision2023}.

The aforementioned studies have delved into the performance of diverse learnable models in distracted driving detection, substantially broadening the horizons of this field. Moreover, relevant research goes beyond these studies. For instance, there is exploration into the application of unsupervised learning \cite{liNewUnsupervisedDeep2022,royUnsupervisedSparseNonnegative2022}, contrastive learning \cite{huDriverAnomalyQuantification2023, yangQuantitativeIdentificationDriver2023}, and vision-language pretraining models \cite{hasanVisionLanguageModelsCan2023}. Recent research has highlighted the significant relevance of digital twins in distracted driving detection tasks \cite{maDriverDigitalTwin2024}. With the emergence of the concept of intelligent cockpit systems \cite{chenMilestonesAutonomousDriving2023}, vision-based distracted driving detection has encountered significant opportunities.

\subsection{Transfer Learning}
Transfer learning (TL) aims to enhance the performance of target learners in target domains by transferring knowledge from different but related source domains \cite{zhuangComprehensiveSurveyTransfer2021}.
Due to its capability to expedite the convergence of CNNs, TL has found widespread adoption, including applications in forecasting residential electric vehicle (EV) charging behavior \cite{forootaniTransferLearningbasedFramework2023}, evaluating driver workload \cite{xiePersonalizedDriverWorkload2020}, and detecting driver distraction \cite{xingDriverActivityRecognition2019,Masood_Detecting2020,beheraLatentBodyPoseGuided2018}.
By applying TL to visual categorization, several common challenges such as view divergence in action recognition tasks and concept drifting in image classification tasks can be effectively addressed \cite{lingshaoTransferLearningVisual2015}.
TL algorithms in visual categorization applications, such as object recognition, image classification, and human action recognition, have demonstrated promising results.
Behera \emph{et al.} \cite{beheraLatentBodyPoseGuided2018} observed that TL is advantageous for cross-dataset performance.
Therefore, despite Baheti \emph{et al.} \cite{Baheti_Computationally2020} reporting limited effects of TL on driver distracted detection, TL is still considered a crucial step in our approach, especially considering the lack of cross-dataset validation in their study.

\subsection{Label Smoothing}
Miscalibration can be worsened by overfitting during training, as minimizing cross-entropy encourages predicted softmax probabilities to align closely with the One-Hot label assignments \cite{liuDevilMarginMarginbased2022}.
Label smoothing (LS) has been used in image classification, language translation, and speech recognition to prevent networks from becoming over-confident \cite{mullerWhenDoesLabel2020}.
It converts deterministic class labels into probability distributions.
For example, applying a weighted average between the uniform distribution and the hard label is used to reduce the overfitting problem during the training of CNNs and further improve classification performance \cite{szegedyRethinkingInceptionArchitecture2016,zhangDelvingDeepLabel2021}.
Relevant methods have also been applied to tasks related to distracted driving detection \cite{alkanatDensityGuidedLabelSmoothing2022,wangFPTFineGrainedDetection2022,liLearningAccurateSpeedy2022,lyuDriverDistractedBehavior2023}.
Lienen \emph{et al.} argued that the use of a smoothed though still precise probability distribution can be questioned from a theoretical perspective.
They proposed a more novel LS, called label relaxation (LR), which deterministic data in terms of a set of probability distributions instead of a single target distribution.
LR leads to a genuine relaxation of the target instead of distortion, thereby reducing the risk of incorporating undesirable bias in the learning process \cite{lienenLabelSmoothingLabel2021}.

\subsection{Muti-Channel Information Fusion}\label{MCIF}
Multi-sensor fusion plays a crucial role in external perception for autonomous vehicles \cite{yaoRadarCameraFusionObject2023}, and equally crucial for driver sensing.
For example, employing multiview camera and multimodal video for distracted driving detection \cite{Wang_100Driver2023,koayContrastiveLearningVideo2023}.
Furthermore, integrating information from multiple backbone networks can enhance detection performance. 
For instance, fusing various local features such as head and hand features \cite{Abouelnaga_Realtime2018,eraqiDriverDistractionIdentification2019}, or combining global features like skeleton and texture \cite{liNovelSpatialTemporalGraph2019,koayOptimallyWeightedImagePoseApproach2021}, or global-local feature fusion \cite{wuPoseawareMultifeatureFusion2021}.
Feature fusion methods commonly involve feature layer concatenation, cascading fully connected layers \cite{wuPoseawareMultifeatureFusion2021}, genetic-weighted ensemble integration \cite{eraqiDriverDistractionIdentification2019}, or MSA module \cite{maRobustMultiviewMultimodal2023}.
However, these methods introduce additional training parameters, leading to potential overfitting issues, especially in scenarios with limited datasets.
Furthermore, a comprehensive score can be obtained by directly summing the prediction scores from multiple channels \cite{liNovelSpatialTemporalGraph2019, Wang_100Driver2023}.
These methods do not introduce additional parameters but susceptible to the influence of noise.

\section{PROPOSED METHOD}\label{section:3}

\subsection{Score-Softmax classifier}\label{Score-Softmax classifier}
\begin{figure}[!t]
	\centering
	\includegraphics[scale=0.42]{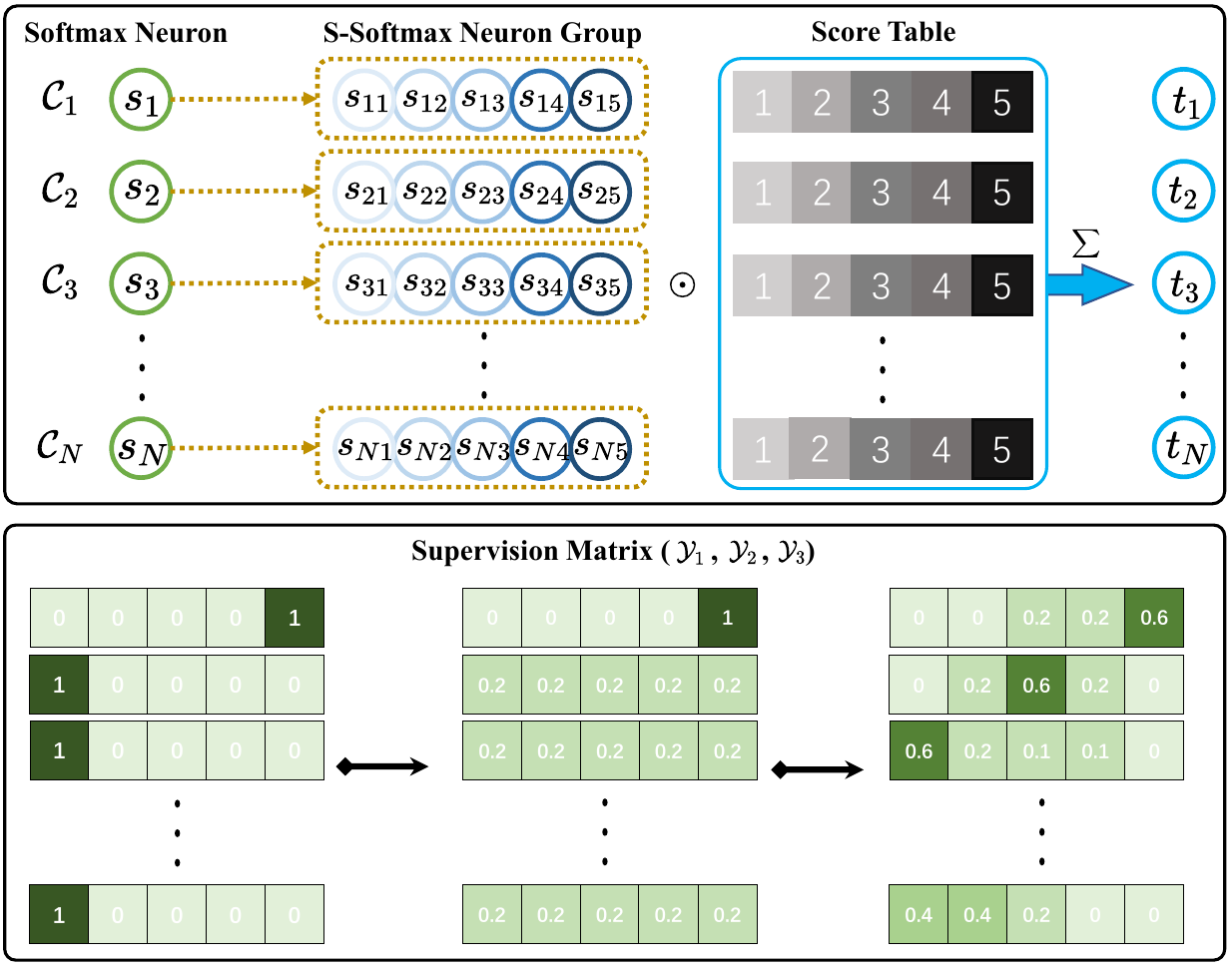}
	\caption{The first row depicts the details of the weighted summation of scores. Blue circles represent individual neuron outputs, each corresponding to different score weights. Each group corresponds to one category. When weighted and summed according to the score table, this output $\mathcal{S}$ yields the score of $\boldsymbol{X}$ for all categories it could belong to. The second row shows three different designs of two-dimensional supervision matrices, from left to right: $\mathcal{Y}_1$, $\mathcal{Y}_2$, and $\mathcal{Y}_3$, where the supervision strength gradually weakens.}
	\label{Figure_classifier}
\end{figure}
\begin{figure*}[!t]
	\centering
	\includegraphics[scale=0.58]{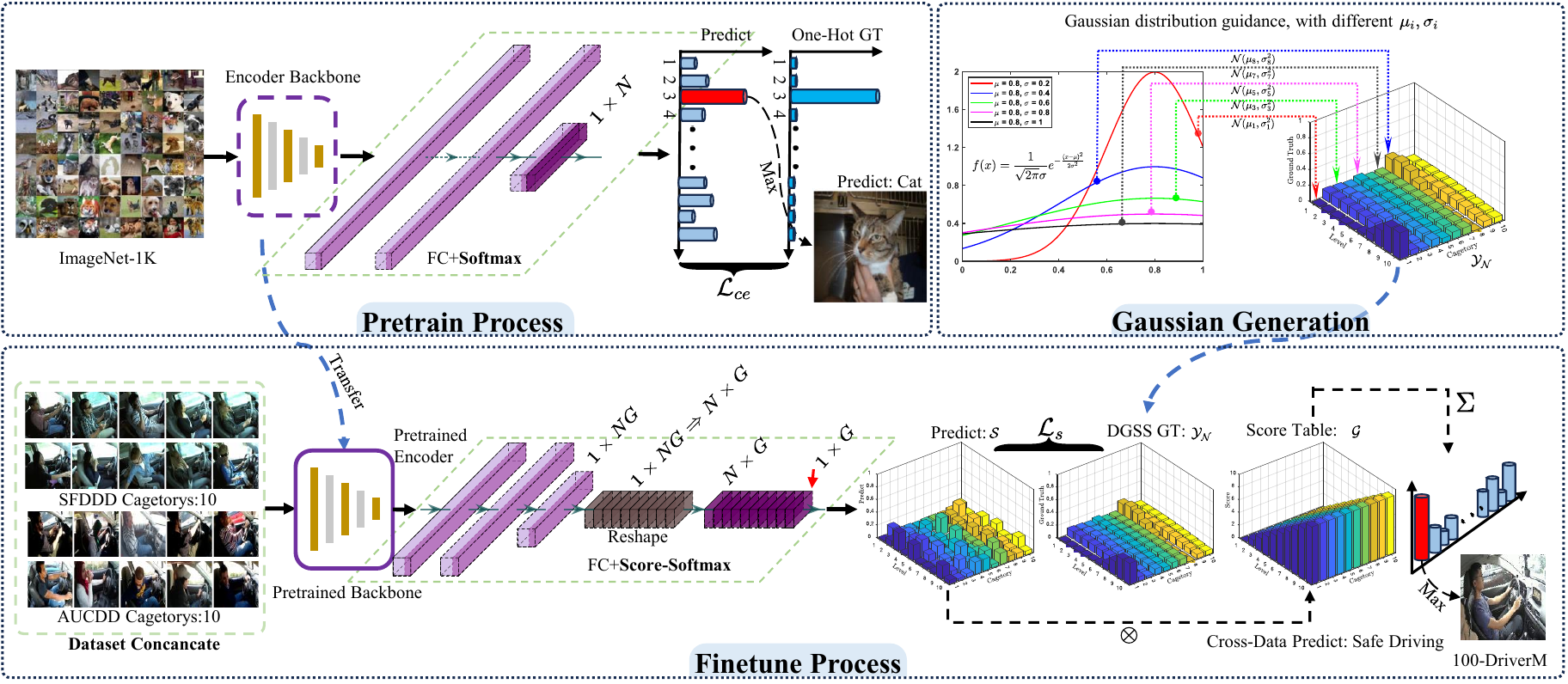}
	\caption{Overview of the training process. The training comprises three main stages: Pretrain Process, Gaussian Generation, and Finetuning Process. In Pretrain Process, we initialize the network with ImageNet dataset training using Softmax classifier and cross-entropy loss $\mathcal{L}_{ce}$. Gaussian Generation involves creating a Gaussian distribution-guided score supervision matrix, a key contribution of our work. In Finetuning Process, we fine-tune the model on the combined distracted driving dataset, transferring pretrained backbone network weights, and employing the S-Softmax classifier with a score loss function $\mathcal{L}_s$. Class scores are computed using weighted summation (represented by $\Sigma$), and weighted product is represented by $\otimes$.}
	\label{Figure_system}
\end{figure*}
The Softmax classifier is currently a primary method for manual distracted driving detection  \cite{Baheti_Detection2018,Baheti_Computationally2020,Bera_Attend2021,Qin_Distracted2022,Li_Driver2022,liLearningAccurateSpeedy2022,Liu_Extremely2023,Mittal_CATCapsNet2023}.
It typically forms the classification module at the end of a CNN alongside fully connected layers. 
Assume $\mathcal{I}_N=\{1,2,\cdots,N\}$, and $\mathcal{C}=\{\mathcal{C}_i|i\in \mathcal{I}_N\}$ denotes all cagetories.
An end-to-end CNN can be formulated as
\begin{equation}
	\label{eq.1}
	\boldsymbol{S} = f_{\text{s}}(f_{\text{fc}}^{\theta}(f_{\text{fe}}^{\theta}(\boldsymbol{X}))),
\end{equation}
where $\boldsymbol{X}$ is the input frame, $\boldsymbol{S}$ is the predicted probability, and $\boldsymbol{S} =\{s_i|i\in \mathcal{I}_N\}$.
The function $f_{\text{fe}}^{\theta}$, $f_{\text{fc}}^{\theta}$, and $f_{\text{s}}$ denote the feature extraction backbone, fully connected layers, and Softmax layer, respectively.
The ${\theta}$ represents the trainable parameters.
Assume $\boldsymbol{P}$ denotes the output of $f_{\text{fc}}^{\theta}$, $\boldsymbol{P} =\{p_i|i\in \mathcal{I}_N\}$,
and $\boldsymbol{S}=f_{\text{s}}(\boldsymbol{P})$.
The $f_{\text{s}}$ can be expressed as
\begin{equation}
	\label{eq.4}
	s_i = \frac{\exp(p_i)}{\sum_{j=1}^{N} \exp(p_j)}.
\end{equation}
The cross-entropy loss $\mathcal{L}_{ce}$ is widely adopted to supervise $f_{\text{s}}$, 
\begin{equation}
	\label{eq_CrossLoss}
	\mathcal{L}_{ce} = -\sum_{i}y_i \log(s_i)\text{,}
\end{equation}
where $y_i$ is the one-hot label.
\modifyRS{This label requires $\sum_{i\in \mathcal{I}_N}y_i=1$, so $y_{i|\mathcal{C}_i}= 1$ is accompanied by $y_{i| \mathcal{C} - \mathcal{C}_i}= 0$, where $\mathcal{C}_i$ denotes the groundtruth category. Current research indicates the combination of Softmax, one-hot labels, and cross-entropy loss explicitly leads to models becoming overly confident \cite{lienenLabelSmoothingLabel2021,zhangDelvingDeepLabel2021,liuDevilMarginMarginbased2022,parkACLSAdaptiveConditional2023}.
However, they overlooked the issues inherent in Softmax itself while focusing on improvements in label smoothing and loss function design. Specifically, the category constraint $\sum_{i\in \mathcal{I}_N}y_i=1$ arising from the one-hot label imposed by Softmax is weakened but never eliminated.
Additionally, label uniqueness, denoted as $y_{i|\forall \boldsymbol{X}\in\mathcal{C}_i}$, is also considered inappropriate. In this scenario, the model's predictions lack uncertainty, whereas uncertainty is deemed crucial \cite{zhangExplainableMachineLearning2022}.}

\modifyRS{In order to eliminate constraint that $\sum_{i\in \mathcal{I}_N}y_i=1$, we expand the $f_{\text{s}}$ to two-dimensional score weighted summation classifier $f_{\text{ss}}$ , called Score-Softmax (S-Softmax).
Fig.\ref{Figure_classifier} shows how
$f_{\text{ss}}$ to remove the constraint by changing the prediction of category probability to the prediction of the weighted distribution of confidence scores.}
We expand $N$ neurons to $N$ neuron groups in the last layer of $f_{\text{fc}}$, with each group consisting of $G$ neurons. 
The $\boldsymbol{P}$ becomes $\{p_i|i\in \mathcal{I}_{N\times G}\}$.
And the $\boldsymbol{S}$ becomes $\mathcal{S}=f_{\text{ss}}(\boldsymbol{P})$, 
\begin{equation}
	\label{eq.7}
	f_{\text{ss}}(\cdot)=f_{\text{s}}^{*}(
		f_{\text{g}}(\cdot))\text{,}
\end{equation}
where $f_{\text{g}}$ denotes grouping operator and $f_{\text{s}}^{*}$ means interior softmax operator in each group.
Meanwhile, we designed a score table  $\mathcal{G}=\mathcal{I}_G$, with score from $1$ to $G$.
Unlike the neurons estimate the probability directly in Softmax, each neuron group estimates  a set of score weights correspond to one category.
The prediction $\mathcal{C}_p$ is determined by $\mathcal{T}$, which is the weighted sum of scores, $\mathcal{T}=\{t_i|i\in \mathcal{I}_N\}$.
\begin{equation}
	\label{eq.8}
	[\mathcal{T}]_{N\times1}=[\mathcal{S}]_{G\times N}^{\mathrm{T}}[{\mathcal{G}]_{G\times 1}} \text{,}
\end{equation}
where $[\cdot]$ denotes matrixization.
The higher score $t_{i}$ indicating greater confidence score of $\boldsymbol{X}\in\mathcal{C}_i$. 
Ultimately, the $\mathcal{C}_p$ is determined by the highest composite evaluation score $t_p$,
\begin{equation}
	\label{eq.9}
	t_p=\underset{t_i \in \mathcal{T}}{\operatorname{topk}} \left(t_i \right)\text{.}
\end{equation}

The groups are independent of each other, and final score of a certain category depends only on the neuron groups related to it, and is independent of other groups.
Diverse supervision matrices can guide the model towards the same objective during learning. 
For example, the three types of supervision matrices shown in Fig.\ref{Figure_classifier}.
Thus, constraints between categories caused by One-Hot encoding are eliminated.
Furthermore, the weighted sum approach can achieve the same effect without performing latent feature augmentation \cite{Peng_TransDARC2022}.

\subsection{Dynamic Gaussian Smoothing Supervision }\label{Expert Gaussian Distribution Score Matrix And Loss}
\modifyRS{As depicted in Fig.\ref{Figure_system}, we initially transfer the pretrained $f_{\text{fe}}^{\theta}$.
Subsequently, we substitute the $f_{\text{fc}}^{\theta}$ and vanilla $f_{\text{s}}$ with new $f_{\text{fc}}^{\theta}$ and $f_{\text{ss}}$.
The $f_{\text{ss}}$ releases the constraints that expounded in Section \ref{Score-Softmax classifier}.} In this section, we will focus on resolving the issue of category label uniqueness.

Inspired by the human scoring mechanism, we assume the CNN plays the role of a panel $\mathcal{M}_M=\{m_l|l\in\mathcal{I}_M\}$, which composed of $M$ scoring experts.
Each scorer $m_l$ is required to write ballot $s_{ijl}$ denotes that the confidence level of $\boldsymbol{X}\in \mathcal{C}_i$ is $j$,
\begin{equation}
	\label{eq.7}
	s_{ijl}=\begin{cases}
		1& \text{ if } \text{they agree } \\
		0& \text{ if } \text{they disagree }
	\end{cases}\text{.}
\end{equation}
For each $\mathcal{C}_i$, each scorer has only one ballot.
$s_{ij}$ denotes the scorer ratio of support the confidence level $\boldsymbol{X}\in \mathcal{C}_i$ is $j$,
\begin{equation}
	\label{eq.8}
	s_{ij} = \frac{1}{M} \sum_{l\in\mathcal{I}_M} s_{ijl}\text{.}
\end{equation}
Ignoring the limited range of $\mathcal{G}$, there is $\mathcal{N}(\lim\limits_{M \to \infty}s_{j|i} ;\mu_{i},\sigma_i)$, where $\mu_{i}$ is mean and $\sigma_i$ is standard deviation.
Our design is to utilize the output $\mathcal{S}$ of the CNN to approximate the voting results of the scoring group,
\begin{equation}
	\label{eq.9}
	\mathcal{S}=\{s_{ij}|i\in \mathcal{I}_N,j\in \mathcal{I}_G,\sum_{j\in \mathcal{I}_G}s_{ij}=1\}\text{.}
\end{equation}
So we require that the supervision matrix $\mathcal{Y}$, which is the same shape as $\mathcal{S}$, should satisfy the same law,
that is
\begin{equation}
	\label{eq.10}
	\mathcal{Y}=\{y_{ij}|i\in \mathcal{I}_N,j\in \mathcal{I}_G,\sum_{j\in \mathcal{I}_G}y_{ij}=1\}\text{.}
\end{equation}
\modifyRS{The $\mathcal{Y}$ does not represent the actual distribution of the votes.; rather, it is generated through hyperparameters to form an ideal marginal Gaussian distribution matrix, which ensures $\mathcal{S}$ forms an appropriate distribution. 
$\mathcal{Y}$ satisfies $\mathcal{N}(y_{j|{i}};\hat{\mu}_{i},\hat{\sigma}_i)$ and it is regenerated after each iteration by controlling $\mu_i$ and $\sigma_i$. 
Thus, the key to supervision lies in designing sensible dynamic range of $\hat{\mu}_i$ and $\hat{\sigma}_i$.}
By dynamically adjusting $\hat{\mu}_i$ and $\hat{\sigma}_i$, a soft constraint is exerted upon the CNN to implement dynamic Gaussian smoothing supervision (DGSS). 
Specifically, adjusting $\hat{\mu}_i$ according to Eq. \ref{eq.11}, $\lambda$ dynamically takes random values in the interval [$\lambda_\text{min}$, $\lambda_\text{max}$], $\lambda_\text{min}\ge 0$, $\lambda_\text{max}\le 1$.
$\hat{\sigma}_i$ is also dynamically sampled from a range [$\sigma_\text{min}$, $\sigma_\text{max}$] in a similar manner.
\begin{equation}
	\label{eq.11}
	\hat{\mu}_i=\lambda G\text{,}
\end{equation}
\begin{figure}[!b]
	\centering
	\includegraphics[scale=0.4]{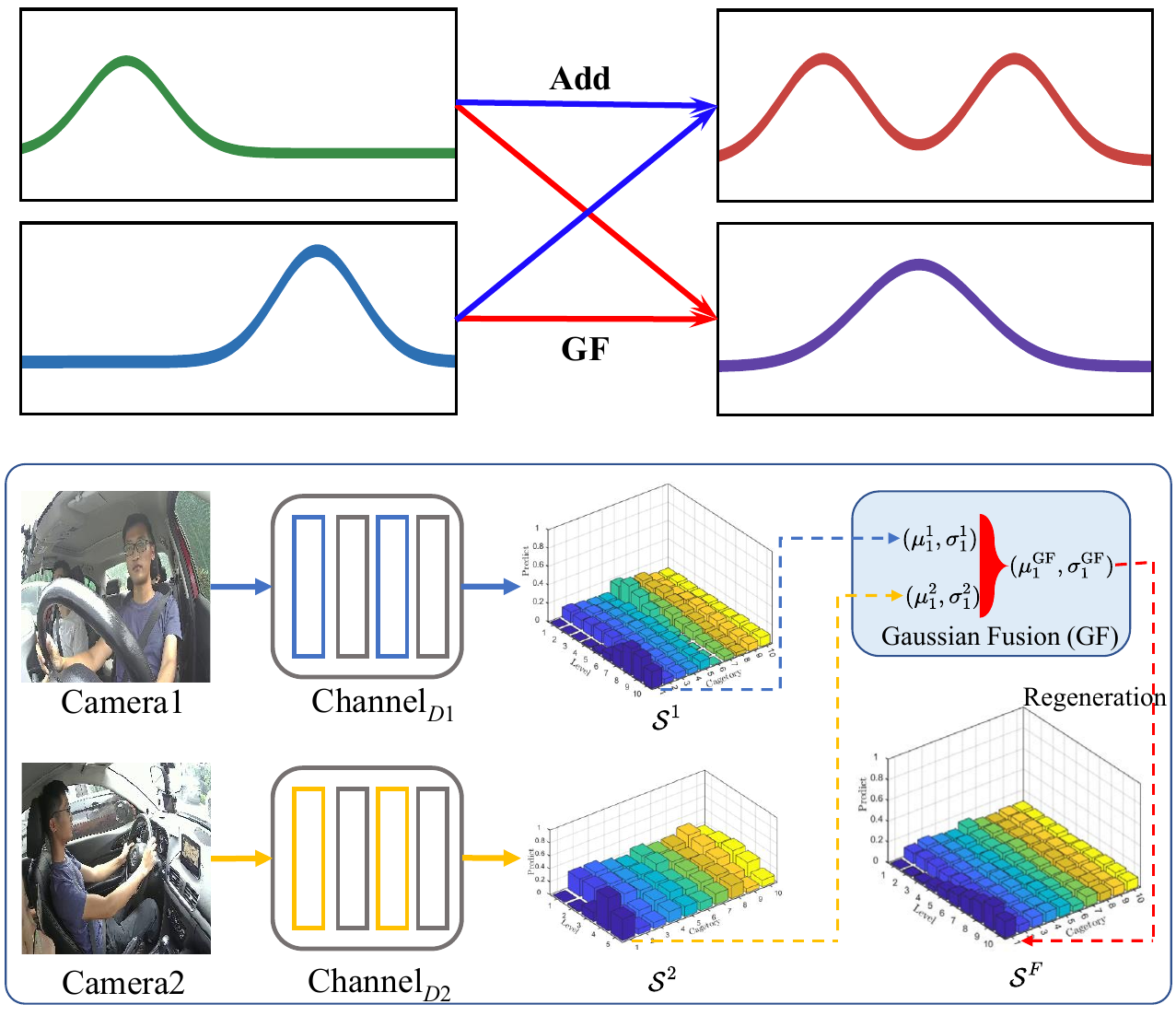}
	\caption{Multi-channel information fusion method based on Gaussian fusion. $\mathcal{S}^1$, $\mathcal{S}^2$ mean the predict score matrix of channl D1 and D1, respectively. $\mathcal{S}^F$ is the fusion score matrix regenerated based on Gaussian distribution.}
	\label{Figure_fusion}
\end{figure}
Additionally, we formulate a loss function denoted as $\mathcal{L}_{s}$, 
\begin{equation}
	\label{eq.13}
	\mathcal{L}_{s}=\|\mathcal{Y}-\mathcal{S}\|_2\text{.}
\end{equation}

\begin{figure}[!h]
	\centering
	\includegraphics[scale=0.5]{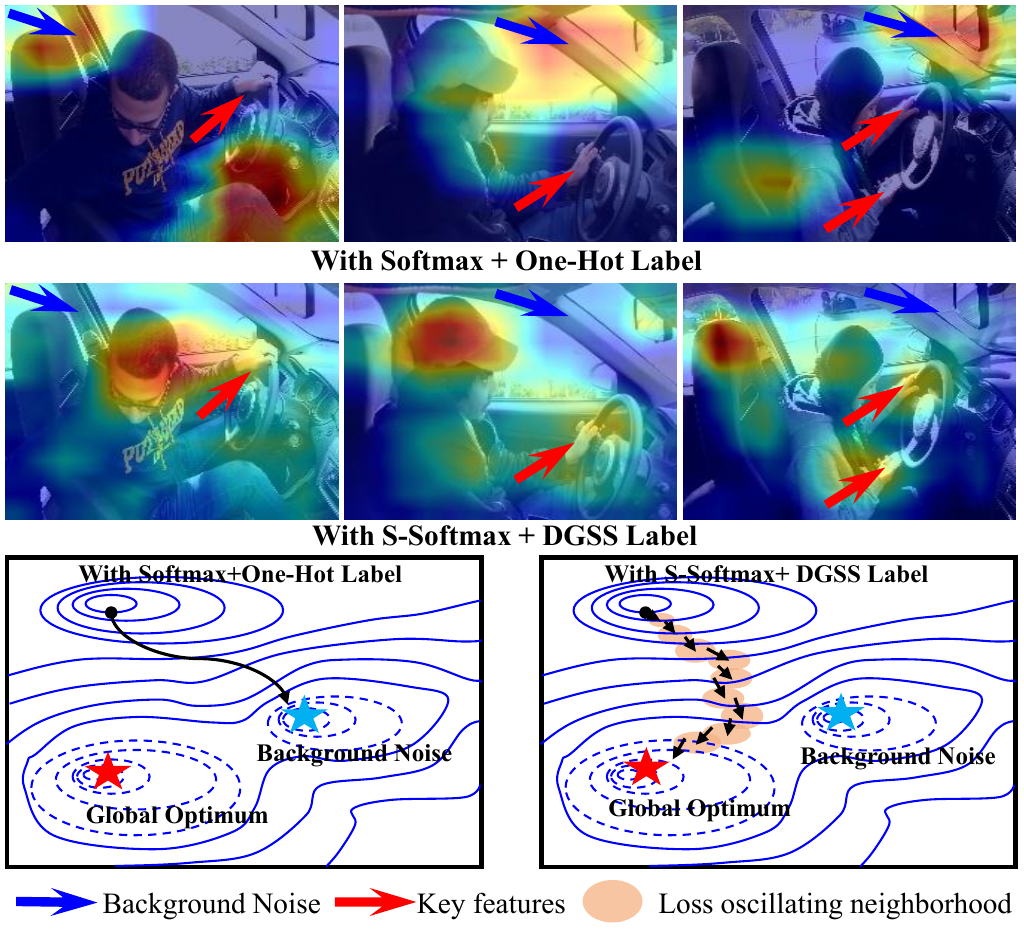}
	\caption{\modifyRS{The first and second rows illustrate attention heatmaps of ResNet18 using Vanilla Softmax with One-Hot labeling, and S-Softmax with DGSS, respectively. These heatmaps are generated through Grad-CAM \cite{selvaraju_grad-cam_2020}. Noise features can create local traps within the solution space. Softmax might lead CNNs to be overly confident, taking shortcuts that could potentially result in falling into traps. To enhance understanding, the brief schematic of the optimization process in the third row vividly demonstrates how DGSS facilitates loss vibration decline, thereby avoiding noise traps.}}
	\label{Figure_optimize}
\end{figure}

Essentially, $\mathcal{Y}$ is a form of Gaussian smoothed label.
But we regenerate the $\mathcal{Y}$ by randomly selecting $\hat \mu_i$ and $\hat \sigma_i$ in each iteration, even for the same sample.
As given in Algorithm~\ref{alg:algorithm1},
this eliminates the intra-class probability unicity.
Therefore, the supervision matrix $\mathcal{Y}$ oscillates within a neighbourhood $\delta(\mathcal{Y})$ centered around true knowledge.
\modifyRS{This increases the uncertainty in model training. Similarly, the Dropout \cite{Dropout} also effectively enhance the generalization ability, which has been explained through uncertainty measurement \cite{zhangExplainableMachineLearning2022}.}
As a result, the loss $\mathcal{L}_s$ likewise oscillates and decreases within the spatial that consist of interconnected $\delta(\mathcal{L}_s)$, as shown in Fig.\ref{Figure_optimize}.
This allows the training to escape shortcuts leading to noise traps.
\begin{algorithm}[t!]
	\caption{S-Softmax Classifier and DGSS.}
	\label{alg:algorithm1}
	\textbf{Require}: Given a dataset $\mathbb{D}=\{(\boldsymbol{X}^{(k)}, y^{(k)})\}_{k=1}^{K}$, where $\boldsymbol{X}^{(k)}\in \mathbb{R}^{\Omega_j}$ represents the $k^{th}$ image, $\Omega_k$ is the spatial image domain. And $y\in\mathcal{I}_N$, it corresponding ground-truth label with $N$ classes. The hyperparameters $\lambda^T$, $\sigma^T$,  $[\lambda_{\text{min}}^F,\lambda_{\text{max}}^F]$, and  $[\sigma_{\text{min}}^F,\sigma_{\text{max}}^F]$.
	\begin{algorithmic}[1] 
		\FOR{epoch $\in$ [1, num\_of\_epoch]}
		\FOR{$(\boldsymbol{X}^{(k)}, y^{(k)})$ in $\mathbb{D}$}
		\FOR{class\_$i$ in $N$}
		\IF{$y^{(k)} = i$}
		\STATE $\hat{\mu}_{i}= \lambda^T\cdot G$
		\STATE $\hat{\sigma}_i= \sigma^T$
		\ELSIF{$y^{(k)} \ne  i$}
		\STATE $\hat{\mu}_{i}= \text{ Random\_sampling}([\lambda_\text{min}^F,\lambda_\text{max}^F])\cdot G$
		\STATE $\hat{\sigma}_i= \text{ Random\_sampling}([\sigma_\text{min}^F,\sigma_\text{max}^F])$
		\ENDIF
		\STATE $y_{j|i}\gets\mathcal{N}(\hat{\mu}_{i},\hat{\sigma}_i)$
		\STATE $\mathcal{Y} \gets y_{j|i}$
		\ENDFOR
		\STATE $\boldsymbol{X}^{*} = {\text{ Augmentor}}(\boldsymbol{X}^{(k)})$
		\STATE $\boldsymbol{P} = f_{\text{fe}} (\boldsymbol{X}^*)$
		\STATE $\mathcal{S} = f_{\rm{ss}} (\boldsymbol{P})$
		\STATE $\mathcal{L}$ = $\mathcal{L}_{\rm s}(\mathcal{Y},\mathcal{S})$ 
		\STATE \textbf{BACKPROP}($\mathcal{L}$)
		\ENDFOR
		\ENDFOR
	\end{algorithmic}
\end{algorithm}
\subsection{Multi-channel feature fusion}\label{Multi-channel feature fusion}
As described in Section \ref{MCIF}, feature fusion is widely adopted in distracted driver detection, primarily for multi-camera and multi-modal fusion. 
The softmax classifier is very convenient for either feature vector summation fusion or concatenation fusion \cite{liNovelSpatialTemporalGraph2019, Wang_100Driver2023}. 
But for S-Softmax, multi-information summation $\mathcal{N}(y_{j|{i}};{\mu}_{i},{\sigma}_i)$ cannot properly represent the fusion result. 
As shown in Fig. \ref{Figure_fusion}, direct summation leads to the appearance of multiple peaks, which is inconsistent with the uniformity of the voting distribution. 
And the $\mathcal{S}_i$ cannot be additively fused when the $G$ is different.
Thus, we propose a score matrix regeneration fusion approach, called Gaussian fusion, based on statistical metric $\mu$, $\sigma$.

Assuming there are $K$ different score tables $\{\mathcal{S}^k|k\in\mathcal{I}_K   \}$, and there is $\mathcal{N}(s_{j|i}^k;\mu_i^k,{\sigma_i^k})$.
They are from $K$ channels and there is no correlation between any pair. 
Regarding the desired fused results $\mathcal{S}^\text{GF}$, there is $\mathcal{N}(s_{ij|i}^\text{GF};\mu_i^\text{GF},{\sigma_i^\text{GF}})$.
However, due to DGSS, the actual $\mu_i^k$ and $\sigma_i^k$ cannot be directly determined, and can only be estimated through the calculation of $\hat{\mu}_i^k$ and $\hat{\sigma}_i^k$ based on $s_{ij}^k$ in $\mathcal{S}^k$. 
The calculations of $\hat{\mu}_i^k$ and $\hat{\sigma}_i^k$ are shown in Eq. \ref{eq.14} Eq. \ref{eq.15}, respectively.
Then the $\mu_i^\text{GF}$ and ${\sigma_i^\text{GF}}$, which are the statistical values of the fused distribution, are caculated by Eq. \ref{eq.16} and Eq. \ref{eq.17}.
\begin{equation}
	\label{eq.14}
	{\mu_i^k}\approx  \sum_{j\in\mathcal{I}_G} j{s_{ij}^k}\text{.}
\end{equation}
\begin{equation}
	\label{eq.15}
	{\sigma_i^k}^2\approx \sum_{j\in\mathcal{I}_G} {(j-\mu_i^k)^2{s_{ij}^k}}\text{.}
\end{equation}
Now we could recover the $\mathcal{S}^\text{GF}$ by $\mathcal{N}(s_{ij|i}^\text{GF};\mu_i^\text{GF},{\sigma_i^\text{GF}})$.
This approach is versatile and can be used for the fusion of any number and type of multi-channel score matrix $\mathcal{S}$, as shown in Fig \ref{Figure_fusion}.
 
\begin{equation}
	\label{eq.16}
	\mu_i^\text{GF}=\frac{1}{K} \sum_{k\in\mathcal{I}_K} {\mu_i^k}\text{,}
\end{equation}
\begin{equation}
	\label{eq.17}
	{\sigma_i^\text{GF}}=\sqrt{\sum_{k\in\mathcal{I}_K} {{(\sigma_i^k)}^2}}\text{.}
\end{equation}

\section{EXPERIMENTS }\label{section:4}

\subsection{Datasets}
\modifyRF{We conducted experiments using five publicly available datasets: SFDDD ($\mathbb{D}_1/\mathbb{D}_1^*/\mathbb{T}_1$) \cite{StateFarm_Distracted}, AUCDD ($\mathbb{D}_2/\mathbb{D}_2^*/\mathbb{T}_2$) \cite{Abouelnaga_Realtime2018}, 100-Driver ($\mathbb{D}_3/\mathbb{D}_3^*/\mathbb{T}_3$) \cite{Wang_100Driver2023}, EZZ2021 ($\mathbb{T}_4$) \cite{ezzouhriRobustDeepLearningBased2021}, and the self-collected dataset HNUDDC1 ($\mathbb{T}_5$). 
The $\mathbb{D}_i$, $\mathbb{D}_i^*$, and $\mathbb{T}_i$ denote vanilla train dataset, augmented train dataset, and test dataset, respectively.
All samples of EZZ2021 and the test subset of HNUDDC1 were used as a common test set.
Behera \emph{et al.} \cite{Behera_Deep2022} provided their manually annotated labels for the vanilla test set of SFDDD. 
Other authors provided the label files for AUCDD, 100-Driver, EZZ2021.
The 100-Driver dataset comprised 22 distinct driving action categories.}
To ensure consistent class labels for cross-dataset experiments, we excluded 10 classes, merged two categories, and referred to this modified dataset as 100-DriverM. 
We employed augmentation techniques on all training inputs, including Gaussian blurring, random scaling, translation and rotation, perspective transformation, color enhancement, and more. 

\subsection{Experiment Setting}
All experiments are performed on a computer featuring an AMD Ryzen 5950X and an Nvidia RTX 4090. 
The operating system is Ubuntu 22.04, and the framework is PyTorch 1.12.0. 
We utilize the Adam optimizer \cite{kingmaAdamMethodStochastic2017} with $\beta_1$=0.9 and $\beta_2$=0.999. 
The training is done from scratch with Xavier initialization parameters. 
This process spanned 30 epochs, with an initial learning rate of 1e-3. 
Learning rate decay occurred at the 2nd and 28th epochs, reducing the rate to one-tenth of the previous stage.
For the fine-tuning process, the backbone pre-trained on ImageNet-1K \cite{dengImageNetLargescaleHierarchical2009} is transferred.
The learning rate is 1e-6, spanning 20 epochs.
The batch size is set to 64, and $L_2$ weight regularization is employed with a weight decay of 1e-3.
\subsection{Model And Evaluation Metrics}
\modifyRF{In ablation experiments, we first selected MobileNetV3-S, ShuffleNetV2, EfficientNetB0, ResNet18, and ResNet50 to explore how S-Softmax affects models with different parameters.
MobileNetV3-S and ShuffleNetV2 are lightweight models with limited learning capacity, while EfficientNetB0, ResNet18, and ResNet50 have more learning capabilities. 
Secondly, we compared the DGSS based on S-Softmax with other label smoothing methods based on Softmax.
Like the Vanilla Label Smoothing (VLS) \cite{szegedyRethinkingInceptionArchitecture2016}, Label Relaxation (LR) \cite{lienenLabelSmoothingLabel2021}, Online Lable Smoothing (OLS) \cite{zhangDelvingDeepLabel2021}, Margin-based Label Smoothing (MbLS) \cite{liuDevilMarginMarginbased2022}, and Adaptive and Conditional Label Smoothing (ACLS) \cite{parkACLSAdaptiveConditional2023}}
During the ablation phase, we visually assessed the performance of our method by examining Receiver Operating Characteristic (ROC) curves and Precision-Recall (P-R) curves. Additionally, we employed t-Distributed Stochastic Neighbor Embedding (t-SNE) visualization to provide an intuitive representation of the classification results.
For the cross-dataset validation experiments on the 100-Driver dataset, we employed the same six models used in \cite{Wang_100Driver2023} to enable a direct comparison with their results. 
\section{RESULT AND DISCUSSION}\label{section:5}
\subsection{Ablation Experiments}
\begin{table}[tb]\scriptsize
	\setlength{\tabcolsep}{3.5pt}
	\caption{Regarding the ablation experiments of $\lambda$ and $\hat{\sigma}_i$. The $\lambda$ is divided into three stages: Low Score (\textbf{LS}): [0, 0.5]; Middle Score (\textbf{MS}): [0.5, 0.75]; and High Score (\textbf{HS}): [0.75, 1]. The $\hat{\sigma}_i$ is also divided into three stages: [0.2, 0.6]; [0.6, 1.0]; and [1.0, 1.4]. $\mathbb{D}_{ij}^*$ Means the Combined Dataset of $\mathbb{D}_i^*$ and $\mathbb{D}_j^*$. The $\mathbb{D}\rightarrow\mathbb{T}$ Means the CNN Trained on $\mathbb{D}$ and Test on $\mathbb{T}$.}
	\centering
	\begin{tabular}{c|ccc|ccc|ccc}
		\bottomrule[1.2pt]
		{\bf{Config}}& \multicolumn{3}{c|}{ $\mathbb{D}_{23}^*$$\rightarrow$ $\mathbb{T}_1$}&\multicolumn{3}{c|}{\cellcolor{lightgray} $\mathbb{D}_{13}^*$$\rightarrow$ $\mathbb{T}_2$}&\multicolumn{3}{c}{$\mathbb{D}_{12}^*$$\rightarrow$ $\mathbb{T}_3$}\\ 
		\hline
		\diagbox[innerwidth=4.2em,height=1.5em]{$\hat{\sigma}_i$}{$\lambda$} &LS&MS&HS&\cellcolor{lightgray}LS&\cellcolor{lightgray}MS&\cellcolor{lightgray}HS&LS&MS&HS\\
		\hline
		$[0.2,0.6]$&79.90&80.74&68.27&\cellcolor{lightgray}62.63&\cellcolor{lightgray}61.62&\cellcolor{lightgray}61.81&70.06&69.47&70.57\\
		$[0.6,1.0]$&81.68&\textbf{82.19}&81.57&\cellcolor{lightgray}\textbf{63.20}&\cellcolor{lightgray}62.15&\cellcolor{lightgray}62.09&\textbf{72.60}&71.43&71.39\\
		$[1.0,1.4]$&81.49&81.63&81.63&\cellcolor{lightgray}62.31&\cellcolor{lightgray}62.23&\cellcolor{lightgray}62.24&71.01&71.45&71.62\\
		\toprule[1.2pt]
	\end{tabular}
	\label{Table_BestSSoftmax}%
\end{table}%
\begin{table*}[!h]\scriptsize
	\setlength{\tabcolsep}{3.5pt}
	\caption{\modifyRF{Ablation Experiments About Classifier and Supervision Matrix, Using the Top-1 Accuracy (\%). $\mathbb{D}_1^*$, $\mathbb{D}_2^*$, $\mathbb{D}_3^*$, $\mathbb{T}_1$, $\mathbb{T}_2$ and $\mathbb{T}_3$ Mean the train set ($\mathbb{D}$) and test set ($\mathbb{T}$) of SFDDD, AUCDD and 100-DriverM. $\mathbb{D}_{ij}^*$ Means the Combined Training Dataset of $\mathbb{D}_i^*$ and $\mathbb{D}_j^*$.
	All S-Softmax Classifier with $G=5$.
	For $\mathcal{N}_\mathcal{Y}$, $\lambda^T=0.8$, $\sigma^T=0.2$, $\lambda^F\in[0,0.5]$ and $\sigma_i^F\in[0.6,1]$. The $\mathbb{D}\rightarrow\mathbb{T}$ Means the CNN Trained on $\mathbb{D}$ and Test on $\mathbb{T}$. And \textbf{Bold Fonts} Mean the Best Result.}}
	\centering
	\begin{tabular}{c|c|c|c|c|c|c|c|c|c}
		\toprule[1.2pt]
		\bf{Cross-Dataset Config}&\multicolumn{1}{c}{ $\mathbb{D}_2^*\rightarrow\mathbb{T}_1$}&\multicolumn{1}{c}{ $\mathbb{D}_3^*\rightarrow\mathbb{T}_1$}& \multicolumn{1}{c|}{ $\mathbb{D}_{23}^*$$\rightarrow$ $\mathbb{T}_1$}&\multicolumn{1}{c}{ $\mathbb{D}_1^*\rightarrow\mathbb{T}_2$}&\multicolumn{1}{c}{ $\mathbb{D}_3^*\rightarrow\mathbb{T}_2$}&\multicolumn{1}{c|}{ $\mathbb{D}_{13}^*$$\rightarrow$ $\mathbb{T}_2$}&\multicolumn{1}{c}{ $\mathbb{D}_1^*\rightarrow\mathbb{T}_3$}&\multicolumn{1}{c}{ $\mathbb{D}_2^*\rightarrow\mathbb{T}_3$}&\multicolumn{1}{c}{$\mathbb{D}_{12}^*$$\rightarrow$ $\mathbb{T}_3$}\\ 
		\bottomrule[1.2pt]
		\bf{Model}& \multicolumn{9}{c}{\textit{\textbf{MobileNetV3-S}}}\\ 
		\hline
		Softmax (w/o TL)&\valueRange{25.50}{1.42}&\valueRange{31.17}{0.70}&\valueRange{44.28}{1.17}&\valueRange{26.42}{2.03}&\valueRange{31.83}{0.33}&\valueRange{45.66}{1.21}&\valueRange{21.03}{3.14}&\valueRange{21.21}{0.26}&\valueRange{36.66}{1.08}\\
		Softmax (w/ TL)&\valueRange{44.00}{0.05}&\valueRange{46.23}{0.07}&\valueRange{64.68}{0.14}&\valueRange{35.43}{1.07}&\valueRange{45.21}{0.05}&\valueRange{52.42}{0.09}&\valueRange{31.55}{0.34}&\valueRange{\textbf{50.64}}{0.08}&\valueRange{54.56}{0.36}\\
		\textbf{S-Softmax} (w/ TL and DGSS, $\mathcal{Y}_\mathcal{N}$)&\valueRange{\textbf{53.83}}{0.08}&\valueRange{\textbf{56.04}}{0.05}&\valueRange{\textbf{67.69}}{0.22}&\valueRange{\textbf{39.18}}{0.09}&\valueRange{\textbf{47.82}}{0.07}&\valueRange{\textbf{55.21}}{0.26}&\valueRange{\textbf{31.57}}{1.35}&\valueRange{46.72}{0.13}&\valueRange{\textbf{60.25}}{0.12}\\
		\hline
		
		\bf{Model}&\multicolumn{9}{c}{\textit{\textbf{ShuffleNetV2}}}\\ 
		\hline
		Softmax (w/o TL)&\valueRange{26.17}{1.10}&\valueRange{21.77}{0.92}&\valueRange{31.70}{1.76}&\valueRange{22.43}{0.77}&\valueRange{26.77}{0.94}&\valueRange{36.01}{1.02}&\valueRange{15.59}{0.78}&\valueRange{21.37}{1.93}&\valueRange{28.76}{2.64}\\
		Softmax (w/ TL)&\valueRange{50.95}{1.11}&\valueRange{50.73}{0.45}&\valueRange{65.93}{0.95}&\valueRange{41.70}{0.97}&\valueRange{47.24}{0.25}&\valueRange{56.74}{0.47}&\valueRange{42.38}{1.42}&\valueRange{47.10}{1.57}&\valueRange{58.19}{0.28}\\
		\textbf{S-Softmax} (w/ TL and DGSS, $\mathcal{Y}_\mathcal{N}$)&\valueRange{\textbf{64.84}}{0.95}&\valueRange{\textbf{61.79}}{0.34}&\valueRange{\textbf{68.26}}{0.28}&\valueRange{\textbf{51.50}}{3.13}&\valueRange{\textbf{48.78}}{1.21}&\valueRange{\textbf{59.29}}{0.23}&\valueRange{\textbf{48.09}}{1.72}&\valueRange{\textbf{58.32}}{0.70}&\valueRange{\textbf{63.86}}{0.44}\\
		\hline

		\bf{Model}&\multicolumn{9}{c}{\textit{\textbf{EfficientNetB0}}}\\ 
		\hline
		Softmax (w/o TL)&\valueRange{34.72}{0.41}&\valueRange{37.14}{1.32}&\valueRange{50.66}{1.52}&\valueRange{28.52}{1.11}&\valueRange{34.04}{0.41}&\valueRange{43.99}{1.70}&\valueRange{21.60}{6.17}&\valueRange{32.17}{1.48}&\valueRange{41.98}{4.63}\\
		Softmax (w/ TL)&\valueRange{58.20}{0.38}&\valueRange{64.92}{0.17}&\valueRange{73.39}{0.32}&\valueRange{42.30}{3.21}&\valueRange{52.05}{0.39}&\valueRange{56.78}{0.23}&\valueRange{45.23}{3.47}&\valueRange{52.02}{0.47}&\valueRange{61.29}{0.28}\\
		\textbf{S-Softmax} (w/ TL and DGSS, $\mathcal{Y}_\mathcal{N}$)&\valueRange{\textbf{63.12}}{0.40}&\valueRange{\textbf{71.36}}{0.35}&\valueRange{\textbf{80.58}}{0.17}&\valueRange{\textbf{43.16}}{1.47}&\valueRange{\textbf{55.71}}{0.14}&\valueRange{\textbf{62.08}}{0.24}&\valueRange{\textbf{48.27}}{0.37}&\valueRange{\textbf{56.04}}{0.33}&\valueRange{\textbf{73.00}}{0.47}\\
		\hline
		
		\bf{Model}&\multicolumn{9}{c}{\textit{\textbf{ResNet18}}}\\ 
		\hline
		Softmax (w/o TL)&\valueRange{43.83}{0.45}&\valueRange{44.46}{2.98}&\valueRange{60.34}{0.96}&\valueRange{35.95}{1.96}&\valueRange{42.30}{1.32}&\valueRange{51.50}{0.82}&\valueRange{32.28}{2.16}&\valueRange{41.16}{0.63}&\valueRange{55.27}{4.84}\\
		Softmax (w/ TL)&\valueRange{60.85}{0.09}&\valueRange{56.89}{0.08}&\valueRange{75.18}{0.39}&\valueRange{48.46}{0.67}&\valueRange{48.20}{0.17}&\valueRange{60.78}{0.11}&\valueRange{49.13}{0.61}&\valueRange{53.80}{0.29}&\valueRange{67.43}{0.29}\\
		\textbf{S-Softmax} (w/ TL and DGSS, $\mathcal{Y}_\mathcal{N}$)&\valueRange{\textbf{65.80}}{0.13}&\valueRange{\textbf{71.41}}{0.18}&\valueRange{\textbf{81.68}}{0.26}&\valueRange{\textbf{48.58}}{0.83}&\valueRange{\textbf{54.23}}{0.08}&\valueRange{\textbf{63.20}}{0.09}&\valueRange{\textbf{55.48}}{1.79}&\valueRange{\textbf{59.54}}{0.05}&\valueRange{\textbf{72.60}}{0.58}\\
		\hline
		
		\bf{Model}&\multicolumn{9}{c}{\textit{\textbf{ResNet50}}}\\ 
		\hline
		Softmax (w/o TL)&\valueRange{33.17}{1.28}&\valueRange{30.11}{2.20}&\valueRange{47.43}{2.27}&\valueRange{26.95}{1.15}&\valueRange{34.13}{1.92}&\valueRange{46.94}{1.89}&\valueRange{17.77}{1.57}&\valueRange{26.70}{2.05}&\valueRange{41.16}{3.71}\\
		Softmax (w/ TL)&\valueRange{61.97}{2.79}&\valueRange{67.84}{1.10}&\valueRange{75.61}{0.31}&\valueRange{41.15}{1.68}&\valueRange{52.42}{0.69}&\valueRange{60.20}{0.77}&\valueRange{\textbf{49.32}}{1.20}&\valueRange{58.85}{3.39}&\valueRange{68.73}{1.51}\\
		\textbf{S-Softmax} (w/ TL and DGSS, $\mathcal{Y}_\mathcal{N}$)&\valueRange{\textbf{65.75}}{1.43}&\valueRange{\textbf{68.13}}{0.72}&\valueRange{\textbf{79.35}}{0.72}&\valueRange{\textbf{47.53}}{0.68}&\valueRange{\textbf{58.05}}{0.37}&\valueRange{\textbf{62.36}}{0.58}&\valueRange{49.22}{1.10}&\valueRange{\textbf{63.34}}{1.64}&\valueRange{\textbf{71.79}}{0.98}\\
		\bottomrule[1.2pt]
	\end{tabular}
	\label{Table_twoDatasetTrain_crossDatasetTest}%
\end{table*}%
\begin{table*}[!h]\scriptsize
	\setlength{\tabcolsep}{3pt}
	\caption{\modifyRF{Comparing the DGSS with other LS methods by Top-1 Accuracy (\%). The Backnone is ResNet18 and S-Softmax Classifier with $G=5$. $\mathbb{D}_1^*$, $\mathbb{D}_2^*$, $\mathbb{D}_3^*$, $\mathbb{T}_1$, $\mathbb{T}_2$ and $\mathbb{T}_3$ Mean the train set ($\mathbb{D}$) and test set ($\mathbb{T}$) of SFDDD, AUCDD and 100-DriverM. $\mathbb{T}_1$ and $\mathbb{T}_2$ Mean the Test Set of EZZ2021 and HNUDDC1. $\mathbb{D}_{ij}^*$ Means the Combined Training Dataset of $\mathbb{D}_i^*$ and $\mathbb{D}_j^*$. $\mathbb{D}\rightarrow\mathbb{T}$ Means the CNN Trained on $\mathbb{D}$ and Test on $\mathbb{T}$. $\mathcal{Y}_1$ and $\mathcal{Y}_2$ are the corresponding matrix in Fig. \ref{Figure_classifier}. For the $\mathcal{N}_\mathcal{Y}$ of DGSS, $\lambda^T=0.8$, $\sigma^T=0.2$, $\lambda^F\in[0,0.5]$ and $\sigma_i^F\in[0.6,1]$. And \textbf{Bold Fonts} Mean the Best Result. \textit{\underline{Italic}} Denotes the Second-best Result.}}
	\centering
	\begin{tabular}{c|c|c|c|c|c|c|c|c|c|c|c|c}
		\toprule[1.2pt]
		\bf{Smoothing}&  $\mathbb{D}_{12}^*$$\rightarrow$ $\mathbb{T}_3$& $\mathbb{D}_{12}^*$$\rightarrow$ $\mathbb{T}_4$& $\mathbb{D}_{12}^*$$\rightarrow$ $\mathbb{T}_5$&\#Avg&$\mathbb{D}_{13}^*$$\rightarrow$ $\mathbb{T}_2$& $\mathbb{D}_{13}^*$$\rightarrow$ $\mathbb{T}_4$& $\mathbb{D}_{13}^*$$\rightarrow$ $\mathbb{T}_5$&\#Avg&$\mathbb{D}_{23}^*$$\rightarrow$$\mathbb{T}_1$& $\mathbb{D}_{23}^*$$\rightarrow$ $\mathbb{T}_4$& $\mathbb{D}_{23}^*$$\rightarrow$ $\mathbb{T}_5$&\#Avg\\ 
		\hline
		Softmax@HL&61.66&\textbf{68.59}&49.82&60.02&57.73&70.65&69.09&65.82&72.91&70.17&62.89&68.66 \\ 
		Softmax@VLS \cite{szegedyRethinkingInceptionArchitecture2016}&64.73&\textit{\underline{66.53}}&53.61&61.62&60.40&\textbf{73.29}&74.78&\textit{\underline{69.49}}&75.54&69.11&69.15&71.27\\ 
		Softmax@LR \cite{lienenLabelSmoothingLabel2021}&60.24&65.57&53.85&59.89&57.22&71.76&71.60&66.86&69.71&66.94&58.84&65.16 \\  
		Softmax@OLS \cite{zhangDelvingDeepLabel2021}&62.92&60.19&55.11&59.41&61.39&\textit{\underline{72.28}}&\textbf{77.52}&\textbf{70.40}&76.05&69.60&\textit{\underline{73.05}}&72.90\\  
		Softmax@MbLS \cite{liuDevilMarginMarginbased2022}&63.28&61.95&\textit{\underline{60.85}}&62.03&59.52&67.46&70.67&65.88&72.23&62.54&65.04&66.60\\ 
		Softmax@ACLS \cite{parkACLSAdaptiveConditional2023}&64.78&59.69&59.11&61.19&61.43&70.82&\textit{\underline{75.50}}&69.25&73.34&67.57&62.32&67.74 \\ 
		S-Softmax@$\mathcal{Y}_1$&64.75&62.72&58.16&61.88&56.92&70.90&68.79&65.54&73.05&75.73&60.82&69.87\\ 
		S-Softmax@$\mathcal{Y}_2$&\textit{\underline{68.66}}&63.00&59.10&\textit{\underline{63.59}}&\textit{\underline{62.44}}&69.83&70.14&67.47&\textit{\underline{77.54}}&\textit{\underline{77.57}}&63.96&\textit{\underline{73.02 }}\\ 
		\textbf{S-Softmax@DGSS (Ours)}&\textbf{73.00}&65.76&\textbf{68.14}&\textbf{68.97}&\textbf{63.20}&69.73&75.08&69.34&\textbf{81.63}&\textbf{78.00}&\textbf{73.81}&\textbf{77.81}\\ 
		\bottomrule[1.2pt]
	\end{tabular}
	\label{Tabel_Ablation_LS}
\end{table*}
\begin{figure*}[!t]
	\centering
	\begin{minipage}[t]{1.0\linewidth}
		\centering
		\begin{tabular}{@{\extracolsep{\fill}}c@{}c@{}c@{}c@{}@{\extracolsep{\fill}}}
			\includegraphics[width=0.25\linewidth]{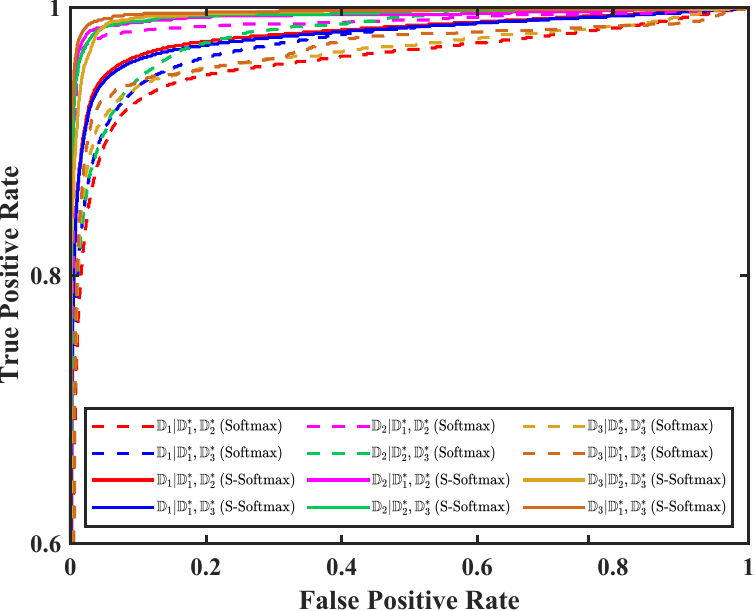}&
			\includegraphics[width=0.25\linewidth]{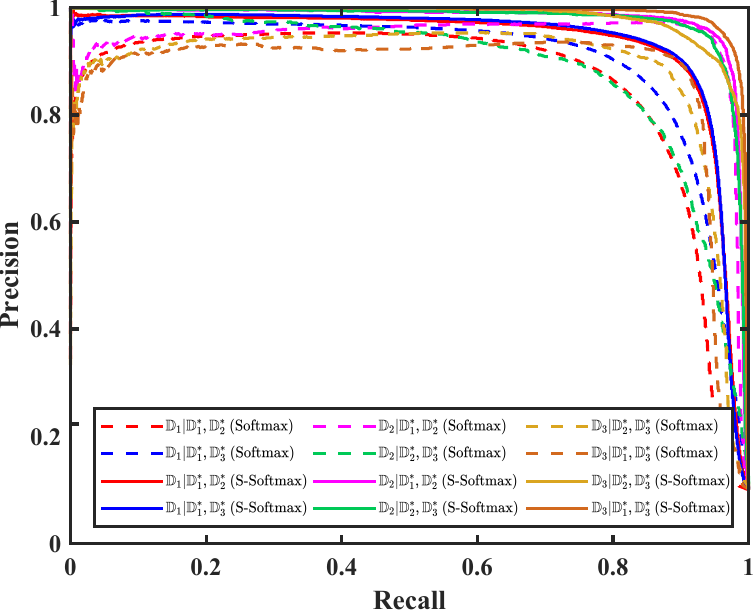}&
			\includegraphics[width=0.25\linewidth]{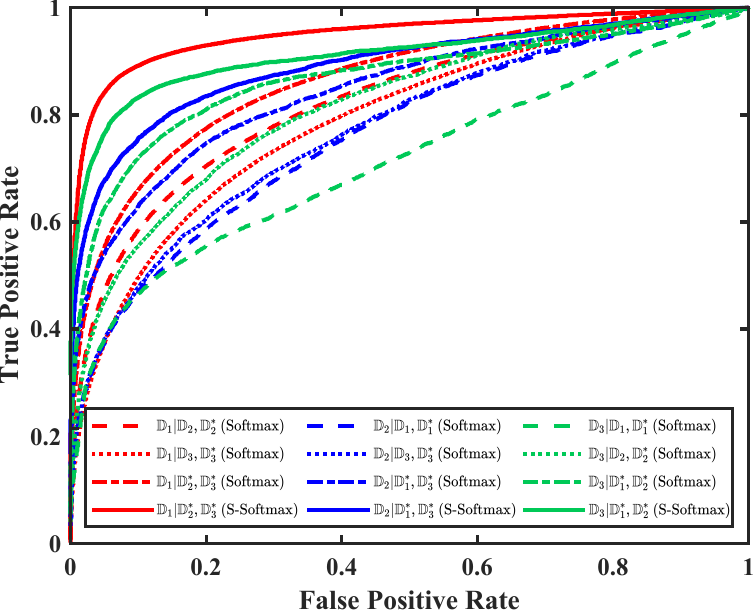}&\includegraphics[width=0.25\linewidth]{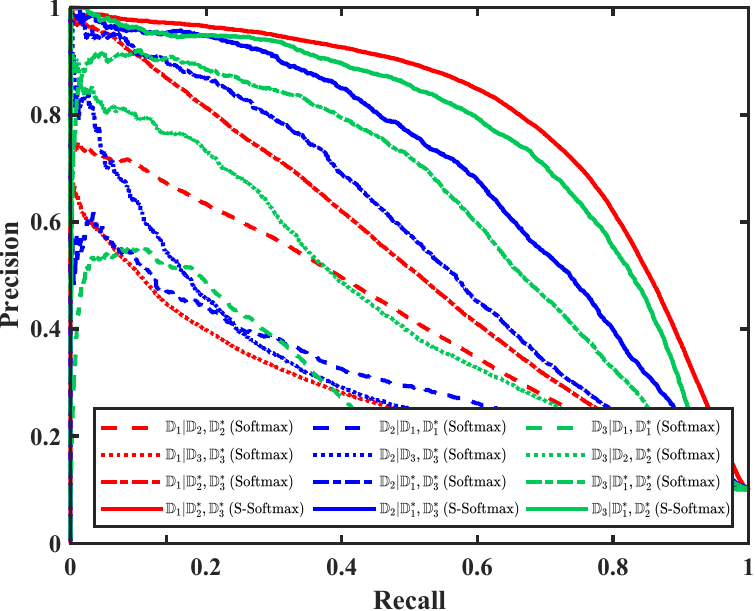}\\
			(a)&(b)&(c)&(d)\\
		\end{tabular}
	\end{minipage}
	\caption{The Receiver Operating Characteristic (ROC) curves and Precision-Recall (P-R) curves based ResNet18 are visualized as follows: (a) and (c) display ROC curves, while (b) and (d) present P-R curves. In (a) and (b), assessments are conducted using the test set associated with the training set, excluding cross-dataset performance evaluations. Conversely, (c) and (d) involve independent test sets, enabling the evaluation of cross-dataset performance. The S-Softmax all with DGSS.}
	\label{Figure_roc_pr}
\end{figure*}
\begin{figure*}[bt]
	\centering
	\includegraphics[scale=0.41]{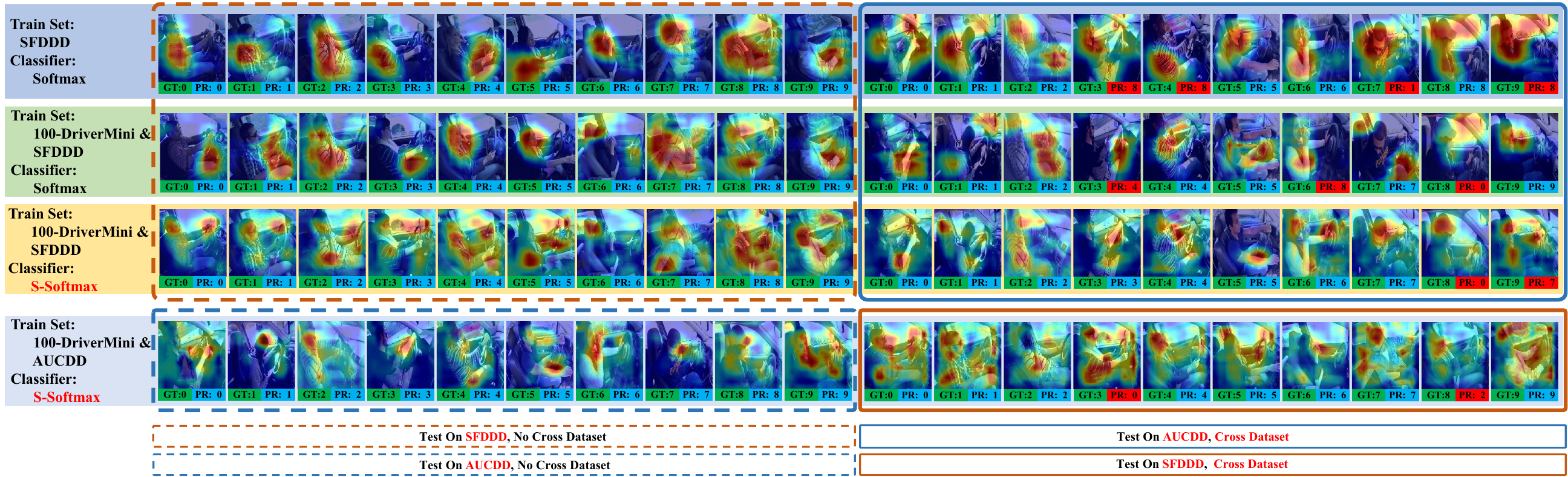}
	\caption{Visualizing the attention regions of ResNet18 using Grad-CAM \cite{selvaraju_grad-cam_2020}. The S-Softmax Classifier with DGSS. The left portion represents non-cross-dataset test results, while the right portion represents cross-dataset test results.}
	\label{Figure_GradCAM}
\end{figure*}

Prior studies suggested that TL is not essential for distracted driving detection \cite{Baheti_Detection2018}. 
This caused subsequent researchers to overlook the importance of TL.
As shown in Table \ref{Table_twoDatasetTrain_crossDatasetTest}, the cross-dataset ablation experiment results demonstrate that transferring CNN models trained on large-scale datasets is indispensable and effectively alleviates overfitting while improves cross-dataset capabilities.
TL resulted in improvements ranging from 5.90\% to 37.73\% across all five models. 
Based on TL, combining multiple datasets also brings significant cross-dataset performance improvements. 
The improvement range is 3.92\% to 20.68\% for the five models.
Based on TL and dataset combination, the proposed S-Softmax classifier and DGSS further enhance cross-dataset metrics.
The adoption of DGSS matrix $\mathcal{Y}_\mathcal{N}$ yields the best improvement results.
For all five models, the range is 0.12\% to 14.52\%.
In fact, the supervision matrix such as $\mathcal{Y}_3$ depicted in Fig. \ref{Figure_classifier} is an instance generated from $\mathcal{Y}_\mathcal{N}$ at one time.
\modifyRF{Compared to the baseline without TL, S-Softmax@DGSS increases the cross-dataset accuracy of MobileNetV3-S, ShuffleNetV2, EfficientNetB0, ResNet18, and ResNet50 by 9.55\% to 21.59\%, 23.28\% to 36.56\%, 18.09\% to 31.02\%, 11.70\% to 21.34\%, and 15.42\% to 31.92\%, respectively.
These results indicate that S-Softmax@DGSS can further enhance the generalization ability based on TL and dataset combination, improving the accuracy of distracted driver monitoring in NDS.}
In all curves of Fig. \ref{Figure_roc_pr} (c) and (d), the ROC and PR curves of S-Softmax@DGSS are above the baseline curves for the corresponding dataset combinations.
This means the improvement in cross-dataset performance is significant.
And the Fig. \ref{Figure_roc_pr} (a) and (b) demonstrate that the it also enhance improve performance on the original test set.

\begin{figure}[bt]
	\centering
	\begin{tabular}{@{\extracolsep{\fill}}c@{}c@{}@{\extracolsep{\fill}}}
		\includegraphics[width=0.5\linewidth]{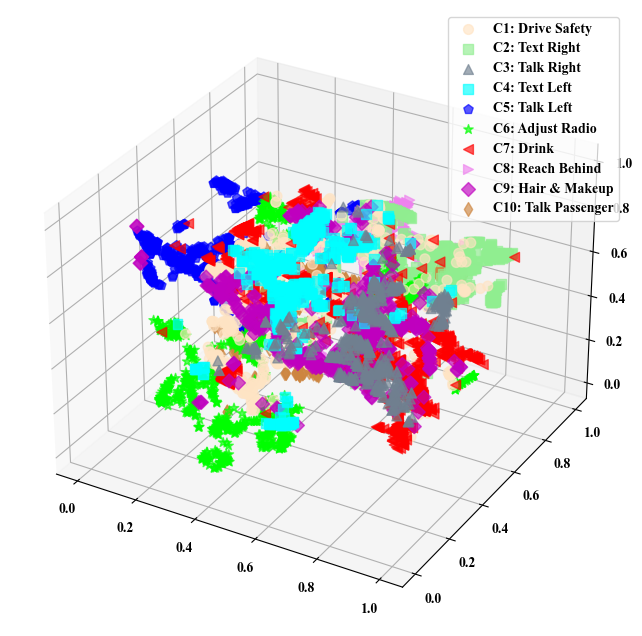}&
		\includegraphics[width=0.5\linewidth]{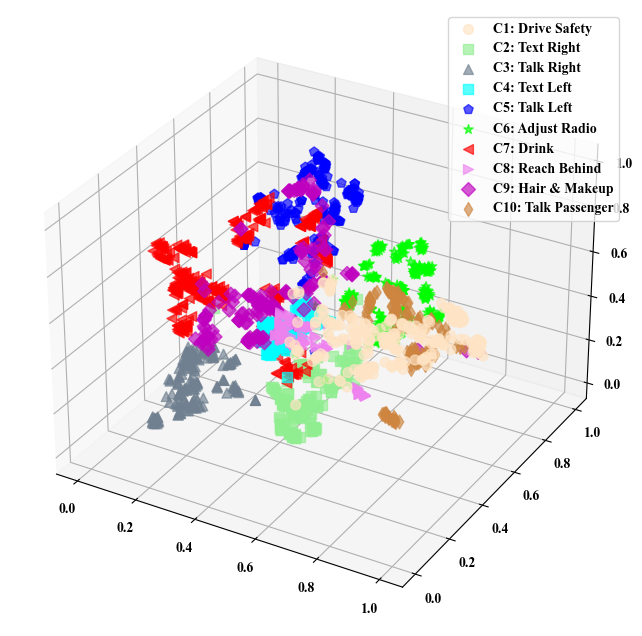}\\
		(a)&(b)
	\end{tabular}
	\caption{\modifyRS{Visualizing the cross-dataset testing results of ResNet18 using t-Distributed Stochastic Neighbor Embedding (t-SNE). The training set is a combined dataset of $\mathbb{D}_{1}$ and  $\mathbb{D}_{2}$, while the test set is $\mathbb{T}_3$. The experiments encompass two distinct training paradigms: (a) results of the Softmax classifier and One-Hot label, (b) results of the S-Softmax classifier and DGSS. These visualizations elucidate that S-Softmax@DGSS improves performance across various categories rather than only specific ones.}}
	\label{Figure_tSNE}
\end{figure}

Fig. \ref{Figure_GradCAM} depicts the feature heatmaps visualized using Grad-CAM \cite{selvaraju_grad-cam_2020} for the trained ResNet18, showing the CNN's focus areas on the tested images. 
A notable improvement is observed where S-Softmax@DGSS results in more dispersed focus areas, with key features relevant to driver behavior receiving increased attention. 
Moreover, high-temperature areas are less concentrated in irrelevant backgrounds. 
Example images in Fig. \ref{Figure_optimize} highlight this improvement, where background noise significantly affects the CNN when using Softmax and One-Hot labels, referred to as the noise trap. This phenomenon is markedly improved with the adoption of S-Softmax and DGSS. 
The loss descent schematic \ref{Figure_optimize} illustrates how DGSS effectively avoids the noise trap. 
\modifyRS{Fig. \ref{Figure_tSNE} indicates that S-Softmax@DGSS shows improvement across all categories, rather than just specific ones. The clustering of samples within the same category is more concentrated, and the distinction between different categories is more pronounced.}

\modifyRF{According to the results in Table \ref{Table_BestSSoftmax}, we set the hyperparameters to $\lambda^F\in [0, 0,5]$ and $\sigma\in[0.2, 0.6]$. Based on this, we compare the proposed S-Softmax and DGSS with other LS methods for cross-dataset performance, as shown in Table \ref{Tabel_Ablation_LS}. 
It indicates that achieving harmonized cross-dataset performance is a highly challenging task, with almost no method achieving optimal performance across all datasets. However, S-Softmax@DGSS achieves the best performance in six out of nine tests, while S-Softmax@$\mathcal{Y}_2$ ranks second in four out of nine tests. No other method achieves the best performance in multiple tests; only OLS ranks first in one test and second in two tests. Although S-Softmax@DGSS does not achieve the highest accuracy in some tests, the difference is small, such as $\mathbb{D}_{12}^*$$\rightarrow$ $\mathbb{T}_4$, $\mathbb{D}_{13}^*$$\rightarrow$ $\mathbb{T}_4$, and $\mathbb{D}_{13}^*$$\rightarrow$ $\mathbb{T}_5$, where the differences are 2.83\%, 3.56\%, and 2.44\%, respectively. When trained on dataset $\mathbb{D_{13}}$, the average difference is only 0.06\%, and it significantly outperforms on the other two datasets on average.}

\begin{figure}[!t]
	\centering
	\includegraphics[scale=0.34]{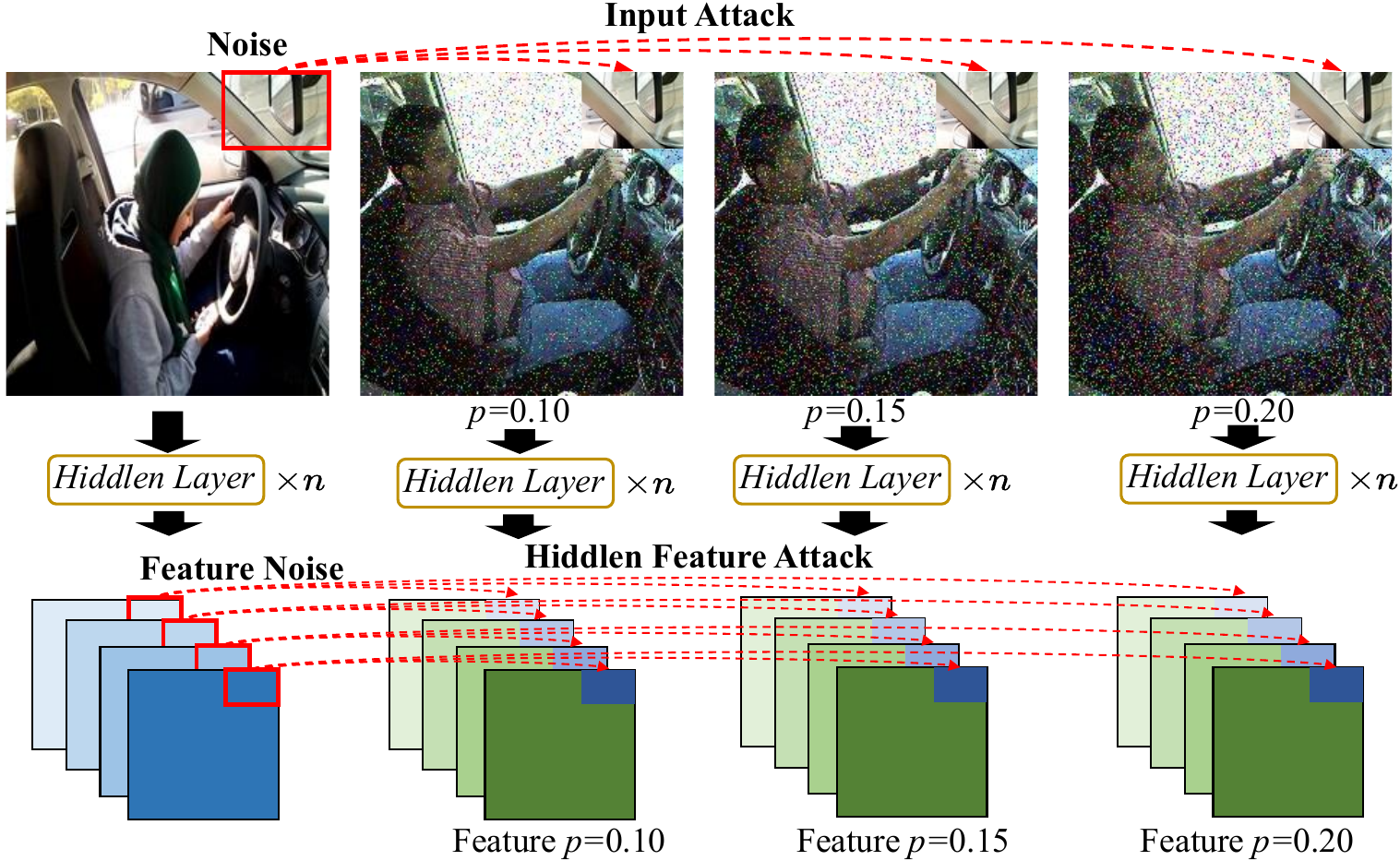}
	\caption{\modifyRS{Diagram of background noise attacks. The noise is selected from the rearview mirror region shown in Fig. \ref{Figure_optimize}. Two noise attack methods are used: input image synthesis (first row) and shallow hidden feature synthesis (second row).}}
	\label{Figure_attack}
\end{figure}
\begin{figure}[t]
	\centering
	\begin{tabular}{@{\extracolsep{\fill}}c@{}c@{}@{\extracolsep{\fill}}}
		\includegraphics[width=0.5\linewidth]{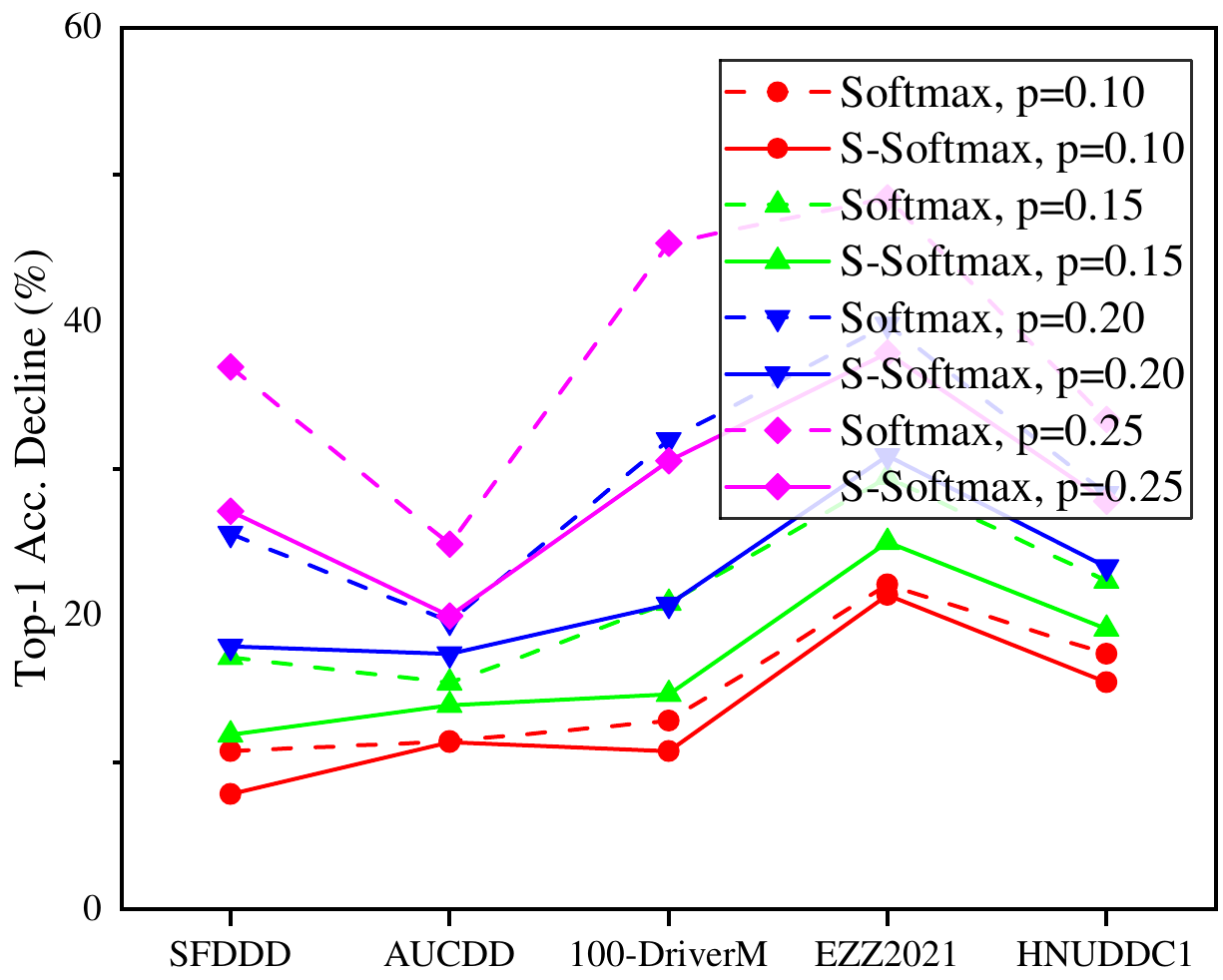}&
		\includegraphics[width=0.5\linewidth]{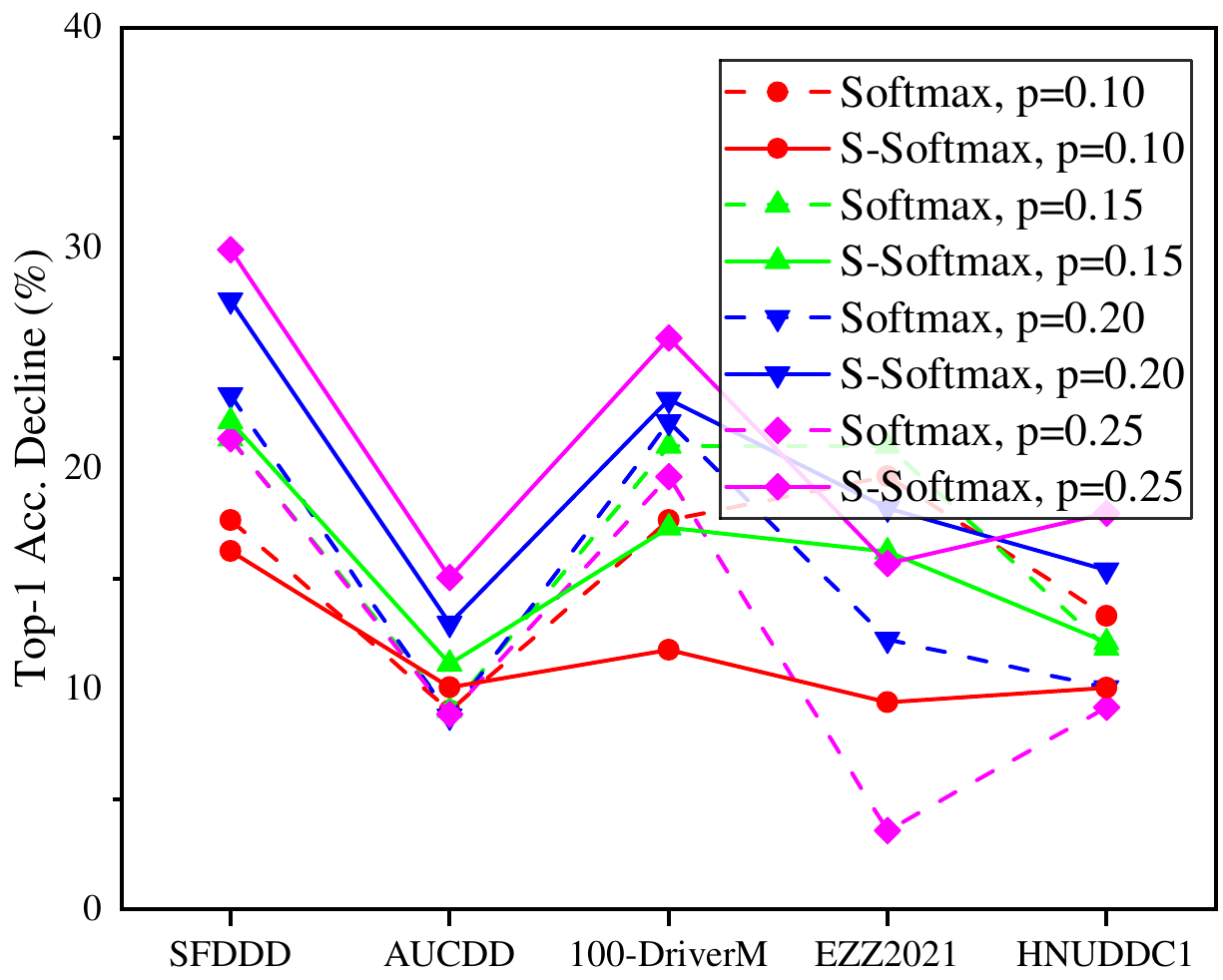}\\
		(a)&(b)
	\end{tabular}
	\begin{tabular}{@{\extracolsep{\fill}}c@{}c@{}@{\extracolsep{\fill}}}
		\includegraphics[width=0.5\linewidth]{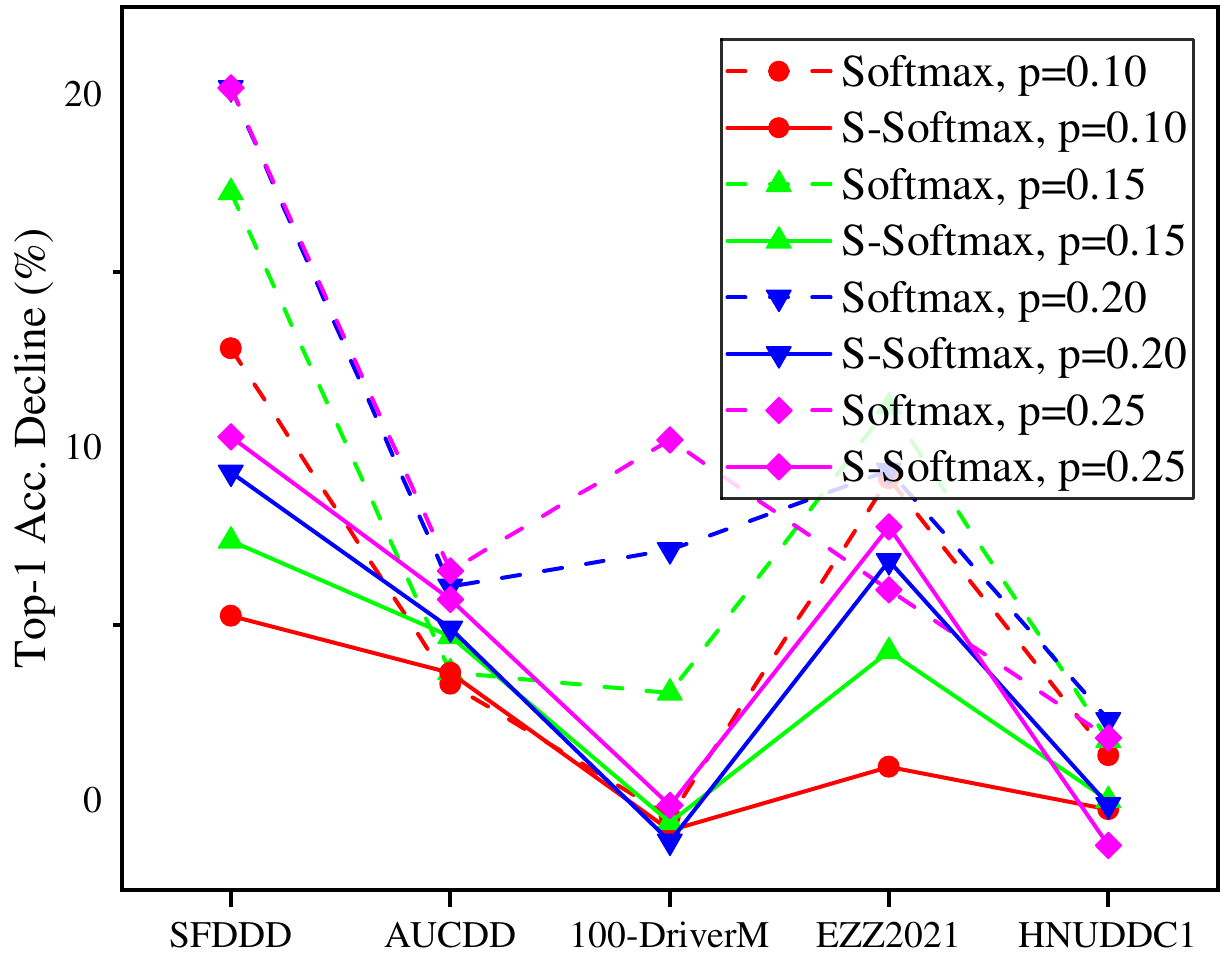}&
		\includegraphics[width=0.5\linewidth]{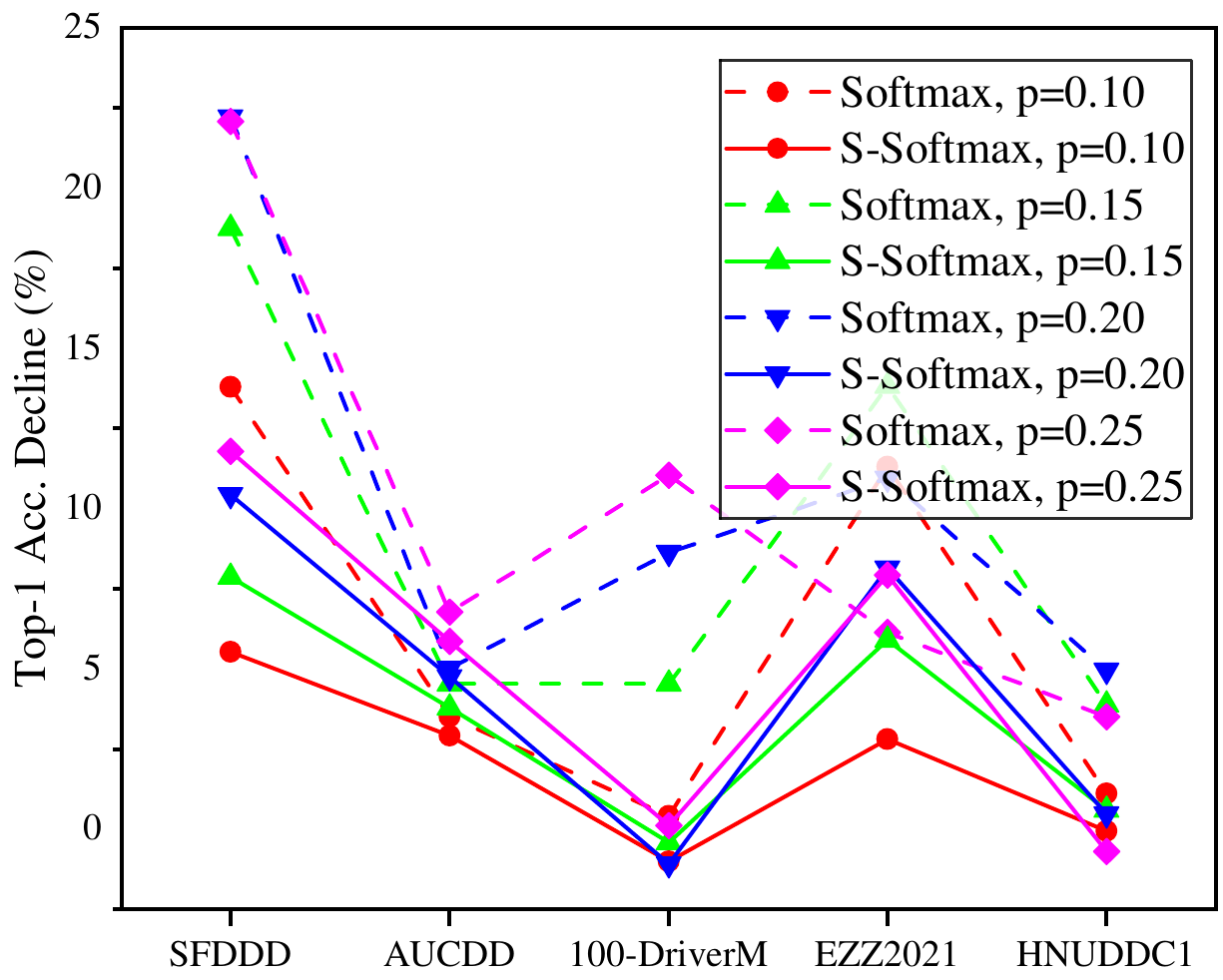}\\
		(c)&(d)
	\end{tabular}
	\caption{\modifyRS{The results of experiments with synthetic distribution shifted data and background noise attacks are presented. (a) shows the results of purely synthetic distribution shift, where the input images are only subjected to impulse noise interference. (b) includes both impulse noise interference and background noise attacks on the input images. (c) and (d) involve impulse noise interference and background noise attacks on hidden features. The background noise in (c) and (d) is extracted from ResNet18 models trained with Softmax and S-Softmax, respectively. The attack is applied after the first convolutional layer of ResNet18.}}
	\label{Figure_attack_result}
\end{figure}

\begin{figure}[b]
	\centering
	\includegraphics[scale=0.5]{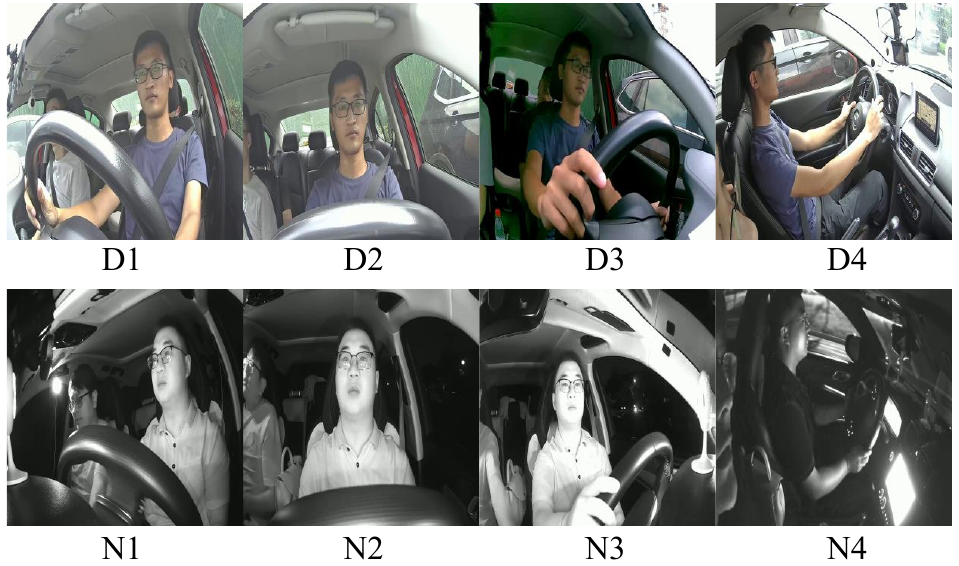}
	\caption{The 100-Driver dataset includes samples captured from various perspectives. Cam1, Cam2, and Cam3 depict frontal views from different angles, while Cam4 presents a side view. The angles between Cam1, Cam2, and Cam3 increase gradually but remain below 30°. In contrast, the angle between Cam4 and the other perspectives is notably larger, exceeding 90°.}
	\label{Figure_100DriverSamples}
\end{figure}

\subsection{Noise Attack Experiments Based on Synthetic Dataset}
\modifyRS{Comparing the accuracy on synthetic distribution-shifted data with the baseline provides an effective measure of model robustness \cite{20212610554416}. Therefore, we tested the model using a test set with impulse noise interference, as shown in Fig. \ref{Figure_attack}. The noise generation probability is denoted by $p$, and the model's resistance to shifted data is evaluated using the Top-1 accuracy decline. In addition to shifted data, background noise can also interfere with the model. Hence, we conducted experiments using conspicuous background noise to attack both the images and features. The results are shown in Fig. \ref{Figure_attack_result}.}

\modifyRS{Fig. \ref{Figure_attack_result} (a) demonstrates that S-Softmax demonstrates stronger resistance to interference compared to Softmax when only impulse noise is added to the input. Furthermore, as the noise intensity increases, S-Softmax withstands stronger interference, as evidenced by the greater difference in performance between the two methods across all five datasets. When background noise attacks are applied to the input images, the situation changes. With weaker impulse noise, such as $p=0.1$ and $p=0.15$, S-Softmax still shows stronger resistance to interference. However, with stronger impulse noise, the accuracy drop is smaller when using Softmax. This is because S-Softmax can focus on key features and avoid background noise traps, which are more noticed by Softmax. Therefore, when impulse noise is weak, the key features remain strong, and the background noise has limited impact on S-Softmax. Conversely, when impulse noise becomes stronger, key features are disrupted. Softmax appears to have ostensible robutness due to the presence of background noise, which actually indicates the model falling into the background noise trap.}

\modifyRS{When background noise attacks the hidden layer features, as shown in Fig. \ref{Figure_attack_result} (c) and (d), S-Softmax demonstrates stronger anti-interference capabilities than Softmax, regardless of the intensity of impulse noise. This indicates that the $7 \times 7$ filter and stride convolution of the first layer of ResNet have a strong filtering ability for background noise in low-dimensional features. Consequently, Softmax's false anti-jamming ability decreases due to the weakening of the background noise trap. At this stage, the key features of high dimension still exist, allowing S-Softmax to maintain strong anti-interference ability. This consistency is also why the trends in Fig. \ref{Figure_attack_result} (c) and (d) are similar. Therefore, S-Softmax improves the overall robustness of the model, whereas the noise trap for Softmax primarily occurs in the first layer of the model. This conclusion is supported by comparing Fig. \ref{Figure_attack_result} (b) with Fig. \ref{Figure_attack_result} (c) or (d).}
\subsection{Cross-dataset Performance Comparison With The State-of-the-art Methods}
\begin{table*}[t]\tiny
	\setlength{\tabcolsep}{4pt}
	\caption{Accuracy (\%) of Cross-View and Cross-Modal on 100-Driver \cite{Wang_100Driver2023}. $\text{D}_i$ Indicates the $i^{th}$ Camera in Day. $\text{N}_i$ Indicates the $i^{th}$ Camera in Night. The Baselines are From \cite{Wang_100Driver2023}, And The \textbf{Bold Datas} Indicate Our Results By Using the S-Softmax and DGSS. The $\textcolor{green}{\uparrow}$ Means Enhanceing And $\textcolor{red}{\downarrow}$ Means Decline. The $\text{D}_i\rightarrow\text{D}_j$ Means the CNN Trained on the Train Set of $\text{D}_i$ and Tested on the Test Set of $\text{D}_j$.}
	\centering
	\begin{tabular}{@{}c|cccc|cccc|cccc|cccc@{}}
		\toprule[1.2pt]
		\bf{Cross Config}&D1$\rightarrow$D2&D1$\rightarrow$D3&D1$\rightarrow$D4&D1$\rightarrow$N1&D2$\rightarrow${D1}&D2$\rightarrow$D3&D2$\rightarrow$D4&D2$\rightarrow$N2&D3$\rightarrow${D1}&D3$\rightarrow$D2&D3$\rightarrow$D4&D3$\rightarrow$N3&D4$\rightarrow${D1}&D4$\rightarrow$D2&D4$\rightarrow$D3&D4$\rightarrow$N4 \\
		\hline \rule{0pt}{8pt}
		ResNet50\cite{Wang_100Driver2023}&50.1&18.4&6.1&16.7&11.2&30.4&6.1&19.2&15.6&31.4&13.1&12.5&5.4&4.1&15.0&33.4\\  
		\textbf{ResNet50(Ours)}&\textbf{56.6}\up&\textbf{27.8}\up&\textbf{4.8}\down&\textbf{57.2}\up&\textbf{47.0}\up&\textbf{47.0}\up&\textbf{8.2}\up&\textbf{47.1}\up&\textbf{26.1}\up&\textbf{46.4}\up&\textbf{13.1}\up&\textbf{41.1}\up&\textbf{3.5}\down&\textbf{4.8}\up&\textbf{17.3}\up&\textbf{60.1}\up\\  
		
		MobileNetV3\cite{Wang_100Driver2023} &48.7&15.0&4.0&21.0&16.6&32.1&2.8&21.7&12.9&25.3&9.1&12.0&4.2&3.5&9.6&14.9\\ 
		\textbf{MobileNetV3(Ours)} &\textbf{55.4}\up&\textbf{22.1}\up&\textbf{5.6}\up&\textbf{45.0}\up&\textbf{25.9}\up&\textbf{29.5}\down&\textbf{5.5}\up&\textbf{35.0}\up&\textbf{18.1}\up&\textbf{36.0}\up&\textbf{11.1}\up&\textbf{23.8}\up&\textbf{4.5}\up&\textbf{4.0}\up&\textbf{11.4}\up&\textbf{44.0}\up\\ 
		
		ShuffleNetV2\cite{Wang_100Driver2023} &44.1&14.7&5.8&5.1&18.9&21.9&5.3&4.8&7.8&26.8&8.8&9.0&3.7&3.4&8.5&3.7\\ 
		\textbf{ShuffleNetV2(Ours)} &\textbf{52.0}\up&\textbf{24.1}\up&\textbf{4.1}&\textbf{44.0}\up&\textbf{34.6}\up&\textbf{37.1}\up&\textbf{5.9}\up&\textbf{39.7}\up&\textbf{16.3}\up&\textbf{36.5}\up&\textbf{10.2}\up&\textbf{29.9}\up&\textbf{4.7}\up&\textbf{3.7}\up&\textbf{11.3}\up&\textbf{38.5}\up\\ 
		
		SqueezeNet\cite{Wang_100Driver2023} &52.1&19.6&5.8&17.1&31.3&38.3&5.4&7.1&14.1&31.8&11.7&6.0&4.9&5.2&11.1&16.4\\ 
		\textbf{SqueezeNet(Ours)} &\textbf{53.7}\up&\textbf{20.1}\up&\textbf{4.8}\down&\textbf{53.2}\up&\textbf{35.8}\up&\textbf{38.7}\up&\textbf{5.8}\up&\textbf{33.3}\up&\textbf{13.1}\down&\textbf{34.3}\up&\textbf{8.7}\down&\textbf{6.8}\up&\textbf{4.2}\down&\textbf{4.0}\down&\textbf{10.3}\down&\textbf{40.9}\up\\ 
		
		EfficienNetB0\cite{Wang_100Driver2023} &51.3&17.3&5.0&13.0&20.7&27.8&4.0&7.9&10.4&28.3&9.0&9.9&5.7&3.8&9.1&21.3\\ 
		\textbf{EfficienNetB0(Ours)} &\textbf{54.4}\up&\textbf{23.6}\up&\textbf{6.5}\up&\textbf{47.8}\up&\textbf{30.9}\up&\textbf{44.8}\up&\textbf{5.9}\up&\textbf{43.1}\up&\textbf{20.7}\up&\textbf{41.3}\up&\textbf{12.4}\up&\textbf{44.7}\up&\textbf{3.1}\down&\textbf{5.1}\up&\textbf{11.3}\up&\textbf{49.0}\up\\
		
		GhostNetV1\cite{Wang_100Driver2023} &48.0&13.1&6.8&12.8&20.5&24.1&4.5&6.3&12.6&25.3&11.8&3.7&3.5&4.0&8.9&5.0\\  
		\textbf{GhostNetV1(Ours)} &\textbf{45.3}\down&\textbf{20.4}\up&\textbf{4.2}&\textbf{31.1}\up&\textbf{24.8}\up&\textbf{28.9}\up&\textbf{4.7}\up&\textbf{35.5}\up&\textbf{17.8}\up&\textbf{37.2}\up&\textbf{11.1}\down&\textbf{16.3}\up&\textbf{4.7}\up&\textbf{4.5}\up&\textbf{9.5}\up&\textbf{30.5}\up\\  
		
		\bottomrule[1.2pt]
	\end{tabular}
	\label{Table_cross_camera_modal}%
\end{table*}%
\begin{table*}[t]\tiny
	\setlength{\tabcolsep}{2.5pt}
	\caption{Accuracy (\%) Of Cross-Vehicle On 100-Driver \cite{Wang_100Driver2023}. $\text{D}_i$ Indicates the $i^{th}$ Camera in Day. \{M, H, A, L\} Represent \{Mazda, Hyundai, Ankai, Lynk\&Co\}. Se Means Sedan. The Baselines Are From \cite{Wang_100Driver2023}, And The \textbf{Bold Datas} Indicate Our Results By Using S-Softmax and DGSS. The $\textcolor{green}{\uparrow}$ Means Enhanceing And $\textcolor{red}{\downarrow}$ Means Decline. The $\text{M}\rightarrow\text{H}$ Means the CNN Trained on the Train Subset of $\text{Mazda}$ and Tested on the Test Subset of $\text{Hyundai}$.}
	\centering
	\begin{tabular}{c|ccccc|ccccc|ccccc|ccccc}
		\toprule[1.2pt]
		\bf{Perspective}&\multicolumn{5}{c|}{D1}&\multicolumn{5}{c|}{D2}&\multicolumn{5}{c|}{D3}&\multicolumn{5}{c}{D4}\\
		\hline \rule{0pt}{8pt}
		{\bf{Cross Dataset Config}}&M$\rightarrow$H&M$\rightarrow$A&M$\rightarrow$L&Se$\rightarrow$SUV&Se$\rightarrow$Van&M$\rightarrow$H&M$\rightarrow$A&M$\rightarrow$L&Se$\rightarrow$SUV&Se$\rightarrow$Van&M$\rightarrow$H&M$\rightarrow$A&M$\rightarrow$L&Se$\rightarrow$SUV&Se$\rightarrow$Van&M$\rightarrow$H&M$\rightarrow$A&M$\rightarrow$L&Se$\rightarrow$SUV&Se$\rightarrow$Van \\  
		\hline \rule{0pt}{8pt}
		ResNet50\cite{Wang_100Driver2023} &27.7&12.3&29.6&36.2&5.2&22.8&0.8&32.6&28.5&1.5&18.9&4.1&29.8&25.4&7.9&32.5&16.8&34.0&42.3&8.0\\   
		\textbf{ResNet50(Ours)} &\textbf{61.4}\up&\textbf{45.8}\up&\textbf{58.8}\up&\textbf{65.6}\up&\textbf{40.7}\up&\textbf{50.3}\up&\textbf{56.4}\up&\textbf{66.0}\up&\textbf{49.5}\up&\textbf{55.7}\up&\textbf{61.6}\up&\textbf{34.5}\up&\textbf{58.0}\up&\textbf{64.9}\up&\textbf{53.0}\up&\textbf{62.7}\up&\textbf{43.1}\up&\textbf{58.0}\up&\textbf{72.3}\up&\textbf{73.7}\up\\ 
		
		MobileNetV3\cite{Wang_100Driver2023}&26.7&11.0&25.9&34.1&7.4&24.1&26.3&32.9&30.7&32.5&21.1&14.5&26.1&23.9&4.8&31.4&4.8&32.5&36.7&0.8\\ 
		\textbf{MobileNetV3(Ours)}&\textbf{57.4}\up&\textbf{21.3}\up&\textbf{50.7}\up&\textbf{66.3}\up&\textbf{24.6}\up&\textbf{43.3}\up&\textbf{43.0}\up&\textbf{56.7}\up&\textbf{45.4}\up&\textbf{32.0}\down&\textbf{52.0}\up&\textbf{17.9}\up&\textbf{42.2}\up&\textbf{62.1}\up&\textbf{36.1}\up&\textbf{57.9}\up&\textbf{11.7}\up&\textbf{57.0}\up&\textbf{65.4}\up&\textbf{59.1}\up\\ 
		
		ShuffleNetV2\cite{Wang_100Driver2023}&30.2&2.1&29.3&28.6&4.6&19.5&0.3&28.6&24.2&1.3&24.8&18.3&27.3&28.8&5.0&31.7&10.3&31.8&38.3&6.0\\ 
		\textbf{ShuffleNetV2(Ours)}&\textbf{56.8}\up&\textbf{32.2}\up&\textbf{52.7}\up&\textbf{60.1}\up&\textbf{21.4}\up&\textbf{41.7}\up&\textbf{42.0}\up&\textbf{57.6}\up&\textbf{43.8}\up&\textbf{41.8}\up&\textbf{51.7}\up&\textbf{10.4}\down&\textbf{47.8}\up&\textbf{58.3}\up&\textbf{16.7}\up&\textbf{57.5}\up&\textbf{56.4}\up&\textbf{52.4}\up&\textbf{59.4}\up&\textbf{46.0}\up\\ 
		
		SqueezeNet\cite{Wang_100Driver2023}&33.4&7.3&35.1&36.1&5.4&26.0&9.5&39.8&31.3&19.1&34.5&25.6&33.6&38.4&24.1&42.0&25.6&38.4&36.4&25.9\\ 
		\textbf{SqueezeNet(Ours)}&\textbf{53.2}\up&\textbf{33.6}\up&\textbf{45.6}\up&\textbf{60.0}\up&\textbf{33.3}\up&\textbf{39.8}\up&\textbf{38.5}\up&\textbf{58.8}\up&\textbf{40.0}\up&\textbf{47.5}\up&\textbf{44.2}\up&\textbf{8.4}\down&\textbf{36.7}\up&\textbf{58.9}\up&\textbf{4.0}\down&\textbf{54.8}\up&\textbf{33.0}\up&\textbf{49.6}\up&\textbf{60.9}\up&\textbf{52.7}\up\\ 
		
		EfficientNetB0\cite{Wang_100Driver2023}&29.2&1.3&32.2&34.1&4.1&26.9&30.3&36.0&30.3&15.9&29.7&11.8&34.0&28.1&19.4&38.9&11.5&39.1&42.5&11.3\\ 
		\textbf{EfficientNetB0(Ours)}&\textbf{61.6}\up&\textbf{26.2}\up&\textbf{55.7}\up&\textbf{68.2}\up&\textbf{19.8}\up&\textbf{48.7}\up&\textbf{50.4}\up&\textbf{61.4}\up&\textbf{51.4}\up&\textbf{33.0}\up&\textbf{56.3}\up&\textbf{14.7}\up&\textbf{45.3}\up&\textbf{64.3}\up&\textbf{15.5}&\textbf{61.5}\up&\textbf{19.9}\up&\textbf{59.4}\up&\textbf{67.8}\up&\textbf{38.3}\up\\ 
		
		GhostNetV1\cite{Wang_100Driver2023}&31.5&8.2&31.3&31.7&2.1&23.1&16.1&33.8&29.5&9.3&18.8&11.9&28.6&27.3&11.8&32.5&11.0&34.7&38.9&0.25\\ 
		\textbf{GhostNetV1(Ours)}&\textbf{53.4}\up&\textbf{25.1}\up&\textbf{44.6}\up&\textbf{58.9}\up&\textbf{30.9}\up&\textbf{34.0}\up&\textbf{40.6}\up&\textbf{51.3}\up&\textbf{42.5}\up&\textbf{39.3}\up&\textbf{46.1}\up&\textbf{11.6}\down&\textbf{39.7}\up&\textbf{56.8}\up&\textbf{25.3}\up&\textbf{47.8}\up&\textbf{22.1}\up&\textbf{45.5}\up&\textbf{58.8}\up&\textbf{44.9}\up\\ 
		\bottomrule[1.2pt]
	\end{tabular}
	\label{Table_cross_vehicle_type}%
\end{table*}%
\begin{table}[tb]\scriptsize
	\setlength{\tabcolsep}{2.2pt}
	\caption{Comparisons of cross-dataset test results between MobileNetV3-S and ResNet18 improved with S-Softmax and DGSS, and the SOTA method for distracted driving detection.  $\mathbb{D}_{ij}^*$ Means the Combined Training Dataset of $\mathbb{D}_i^*$ and $\mathbb{D}_j^*$. The $\mathbb{D}\rightarrow\mathbb{T}$ Means the CNN Trained on $\mathbb{D}$ and Test on $\mathbb{T}$.}
	\centering
	\begin{tabular}{c|c|c|c|c|c}
		\toprule[1.2pt]
		\bf{Model}&\#Param&FLOPs& \multicolumn{1}{c|}{ $\mathbb{D}_{23}^*\rightarrow\mathbb{T}_{1}$}&\multicolumn{1}{c|}{ $\mathbb{D}_{13}^*\rightarrow\mathbb{T}_{2}$}&\multicolumn{1}{c}{$\mathbb{D}_{12}^*\rightarrow\mathbb{T}_{3}$}\\ 
		\bottomrule[1.2pt]
		
		MobileVGG\cite{Baheti_Computationally2020}&1.97M&1.20G&54.09&54.35&55.58\\
		NguyenCNN\cite{Duy-LinhNguyen_Driver2022}&0.46M&1.41G&63.28&54.97&53.05\\
		OLCMNet\cite{Li_Driver2022}&10.18M&3.42G&58.04&51.24&55.35\\
		ELDDR-NAS-KT(S)\cite{Liu_Extremely2023}&\textbf{0.42M}&2.25G&39.25&44.84&46.42\\
		SL-DDBD\cite{zhangNovelDriverDistraction2023}&195.27M&35.81G&80.38&61.28&72.84\\
		MobileNetV3-S(Ours)&1.53M&\textbf{0.06G}&67.69&55.21&60.25\\
		ShuffleNetV2(Ours)&1.26M&0.15G&68.26&59.29&63.86\\
		EfficientNetB0(Ours)&4.02M&0.4G&80.58&62.08&\textbf{73.00}\\
		ResNet18(Ours)&11.18M&1.82G&\textbf{81.68}&\textbf{63.20}&72.60\\
		ResNet50(Ours)&23.53M&4.11G&79.35&62.36&71.79\\

		\bottomrule[1.2pt]
	\end{tabular}
	\label{Table_StateOfTheArt}%
\end{table}%
The real-world performance of distracted driving detection algorithms in NDS is influenced by various complex factors, such as different viewpoint, vehicle and modal.
Recently, the 100-Driver dataset has been specifically collected for driver monitoring videos with cross-viewpoint, cross-vehicle, and cross-modal settings, as shown in Fig. \ref{Figure_100DriverSamples}. 
The authors provided benchmarks for six end-to-end CNNs, which are ResNet50, MobileNetV3-L, ShuffleNetV2, SquueezeNet, EfficientNetB0 and GhostNetV1, in these three cross-X settings \cite{Wang_100Driver2023}.
We adopt the same cross-dataset settings as in \cite{Wang_100Driver2023} and modify the model's classifier to S-Softmax. 
During training, we employ DGSS to validate the effectiveness of the proposed method.
Table \ref{Table_cross_camera_modal} presents the cross-viewpoint and cross-modal results, and Table \ref{Table_cross_vehicle_type} lists the cross-vehicle results.

As shown in Fig. \ref{Figure_100DriverSamples}, the 100-Driver dataset comprises data from three frontal viewpoints, denoted as D1, D2, and D3, and one side viewpoint, denoted as D4. 
Taking D1 as the reference, the angles between D2, D3, and D1 gradually increase. 
Additionally, the angle between D4 and D3 even exceeds the angle between any two frontal viewpoints.
The experiments demonstrate that the proposed method significantly improves the accuracy of cross-viewpoint testing.
Especially for testing between the two adjacent viewpoints of group (D1, D2) and (D2, D3).
The performance improvement of ResNet50 on D2$\rightarrow$D1 reached an astonishing 35.8\%.
For the larger disparity between the frontal views D1 and D3, the highest improvement in cross-dataset testing reached up to 10.3\%.
However, for D4, due to its significant difference from D1, D2, and D4, S-Softmax and DGSS are unable to address this issue.
This is understandable since there are significant changes in key features of driver behavior in D4. Our proposed method primarily reduces the impact of background noise rather than enhancing the ability of CNN to capture entirely different features.
Contrastive Language-Image Pretraining (CLIP) might be able to address this issue due to its added linguistic descriptions \cite{radfordLearningTransferableVisual}.
For cross-modal testing, our method leads to significant improvements for all models across the four viewpoints.
The largest improvement is in the Cam1 viewpoint for ResNet18, reaching 50.5\%.
Light enhances the driver's texture features in daytime, but also highlights background noise of highly reflective surfaces at night, such as the rearview mirror in D4 and the central control screen in N4 in Fig. \ref{Figure_100DriverSamples}. 
These overwhelming features can mislead CNNs, as shown in Fig. \ref{Figure_GradCAM}.
Our method significantly alleviates the issue of transitioning between daytime and nighttime driving environments in NDS.

The results of cross-vehicle validation in Table. \ref{Table_cross_vehicle_type} further demonstrate the significant advantage of the proposed S-Softmax and DGSS.
There are slight differences in the internal structure and some equipment among different brands of vehicles, like Mazda (M), Hyundai (H), Ankai (A), and Lynk\&Co (L) in 100-Driver.
The control area may use traditional buttons or a touchscreen, and variations in camera installation positions arise due to differences in vehicle body structure.
These differences may be greater among different vehiculary types, such as the sedan (Se), sport utility vehicle (SUV) and van.
These differences easily become new traps due to limited datasets, and what is most concerning is that these issues are widespread in NDS.
S-Softmax and DGSS significantly improve this issue, especially for the D4 viewpoint.
The highest improvements for M$\rightarrow$H, M$\rightarrow$A, M$\rightarrow$L, Se$\rightarrow$SUV, and Se$\rightarrow$Van are 27.2\%, 46.1\%, 24.5\%, 30.0\%, and 65.7\%, respectively.
This may be because the D4 viewpoint has more background noise, while the other three primarily focus on the driver.
Regardless, the improvements brought about by our proposed method are comprehensive, indicating its practical value.

Furthermore, we also conducted cross-dataset performance comparisons with the recently proposed state-of-the-art (SOTA) model for distracted driving, and the results are shown in Table \ref{Table_StateOfTheArt}.
MobileVGG, NguyenCNN, and OLCMNet achieve an accuracy of over 50\% in all three types of cross-dataset testing, but OLCMNet has excessively high parameter counts (\#Params) and multiply-accumulate operations (MACs).
While the most lightweight student network, ELDDR-NAS-KT(S), obtained through network architecture search and knowledge transfer, has only 0.42M parameters, but the cross-dataset performance is uninvolved in \cite{Liu_Extremely2023}. 
However, its drawbacks are fully exposed in Table \ref{Table_StateOfTheArt}.
Even though it has only 0.06 GMACs, the MobileNetV3-S improved by S-Softmax and DGSS outperforms all the aforementioned models.
The gain for ResNet18 is even more pronounced, surpassing the latest SL-DDBD, which consists of multiple Swin Transformer Blocks. 
This block is an architecture based on Multi-Head Self-Attention (MSA).
Unlike convolutional filters capture local features in CNNs, MSA prevents the negative impact of background noise through self-attention mechanisms across different regions. 
However, this approach comes with the cost of a large number of parameters and MACs.
S-Softmax and DGSS enable the purely convolutional ResNet18 to surpass SL-DDBD, greatly reducing the model cost.
\begin{figure*}[!t]
	\centering
	\begin{tabular}{@{\extracolsep{\fill}}c@{}c@{}c@{}c@{}@{\extracolsep{\fill}}}
		\includegraphics[width=0.25\linewidth]{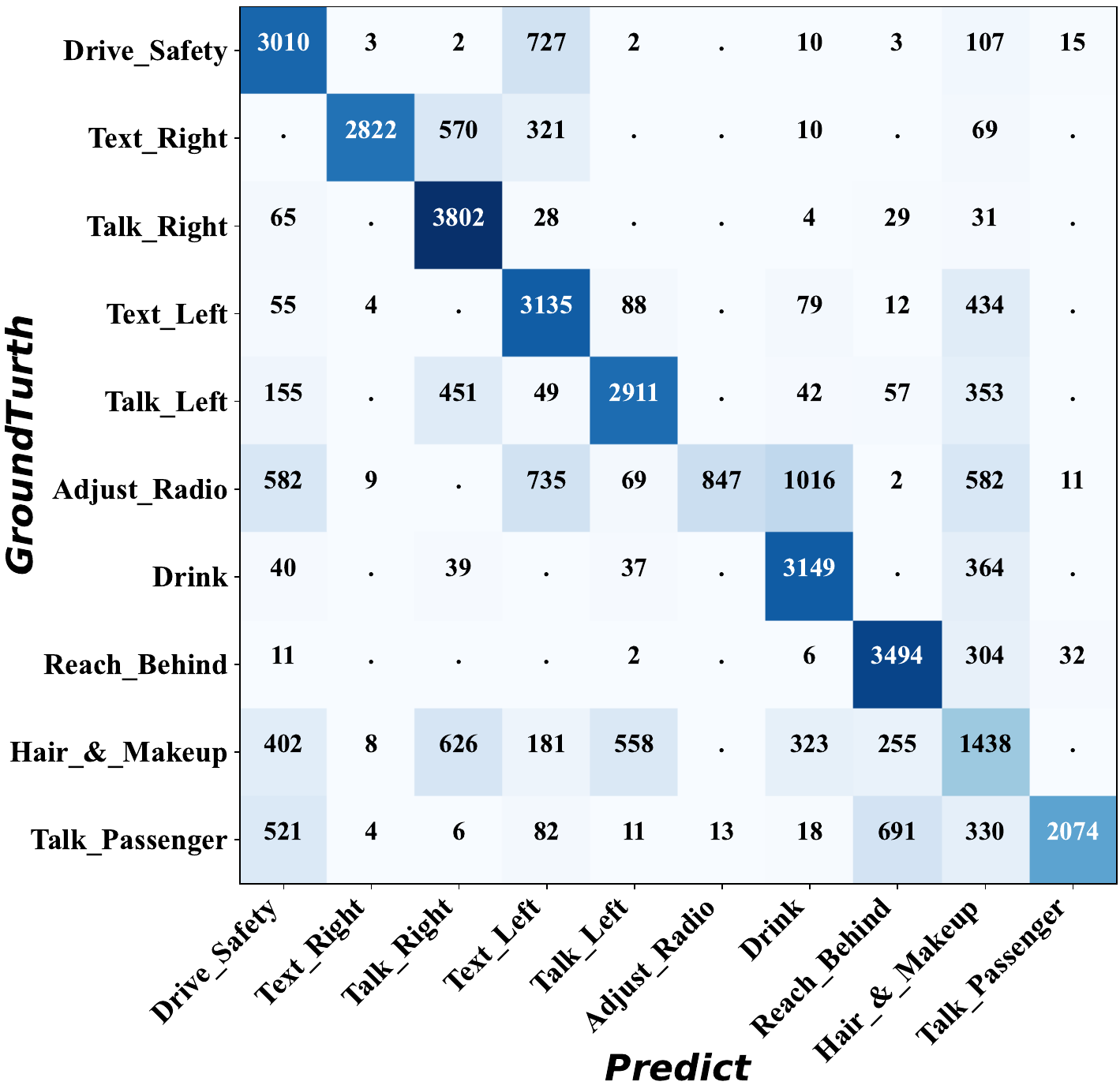}&
		\includegraphics[width=0.25\linewidth]{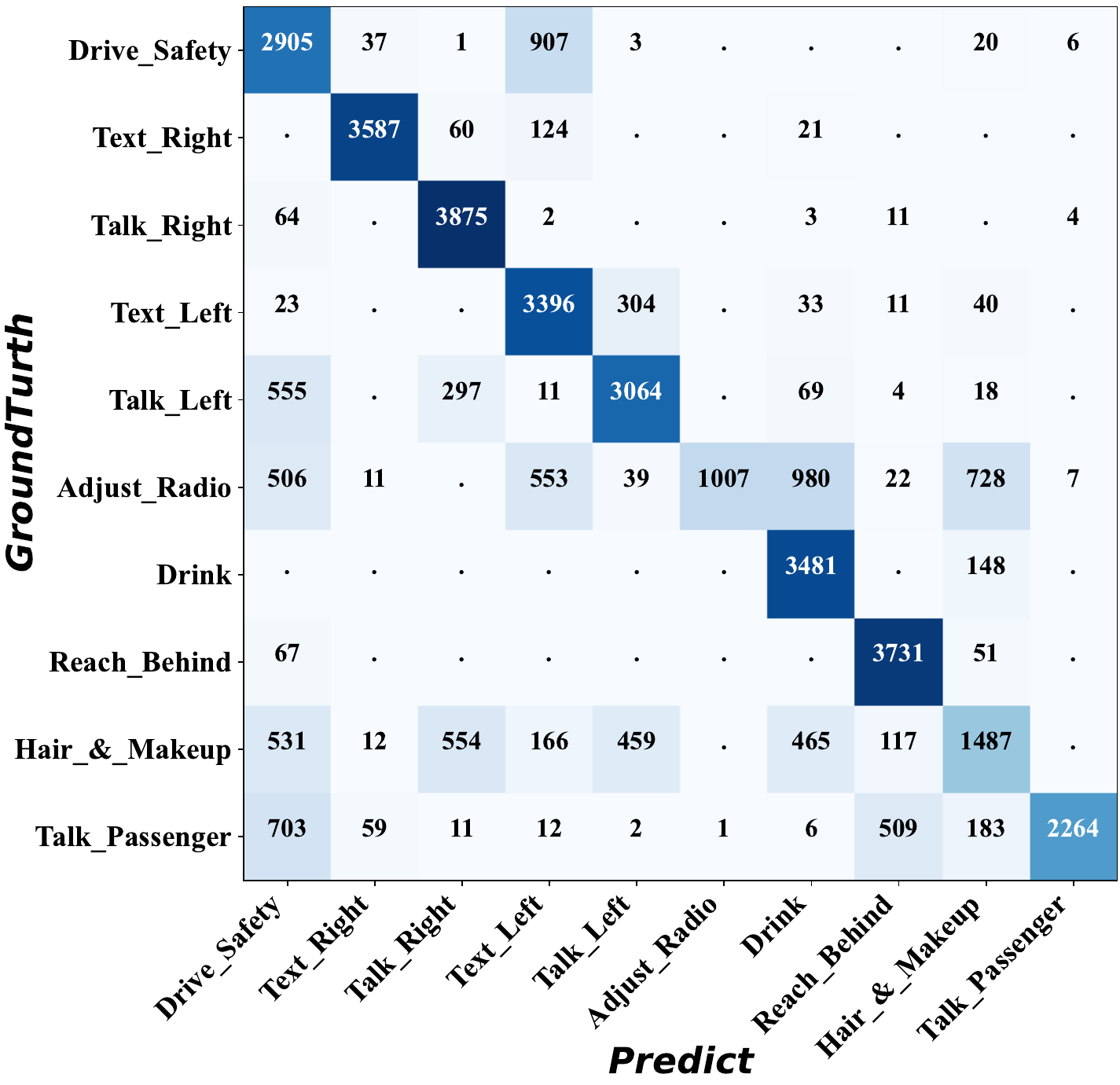}&
		\includegraphics[width=0.25\linewidth]{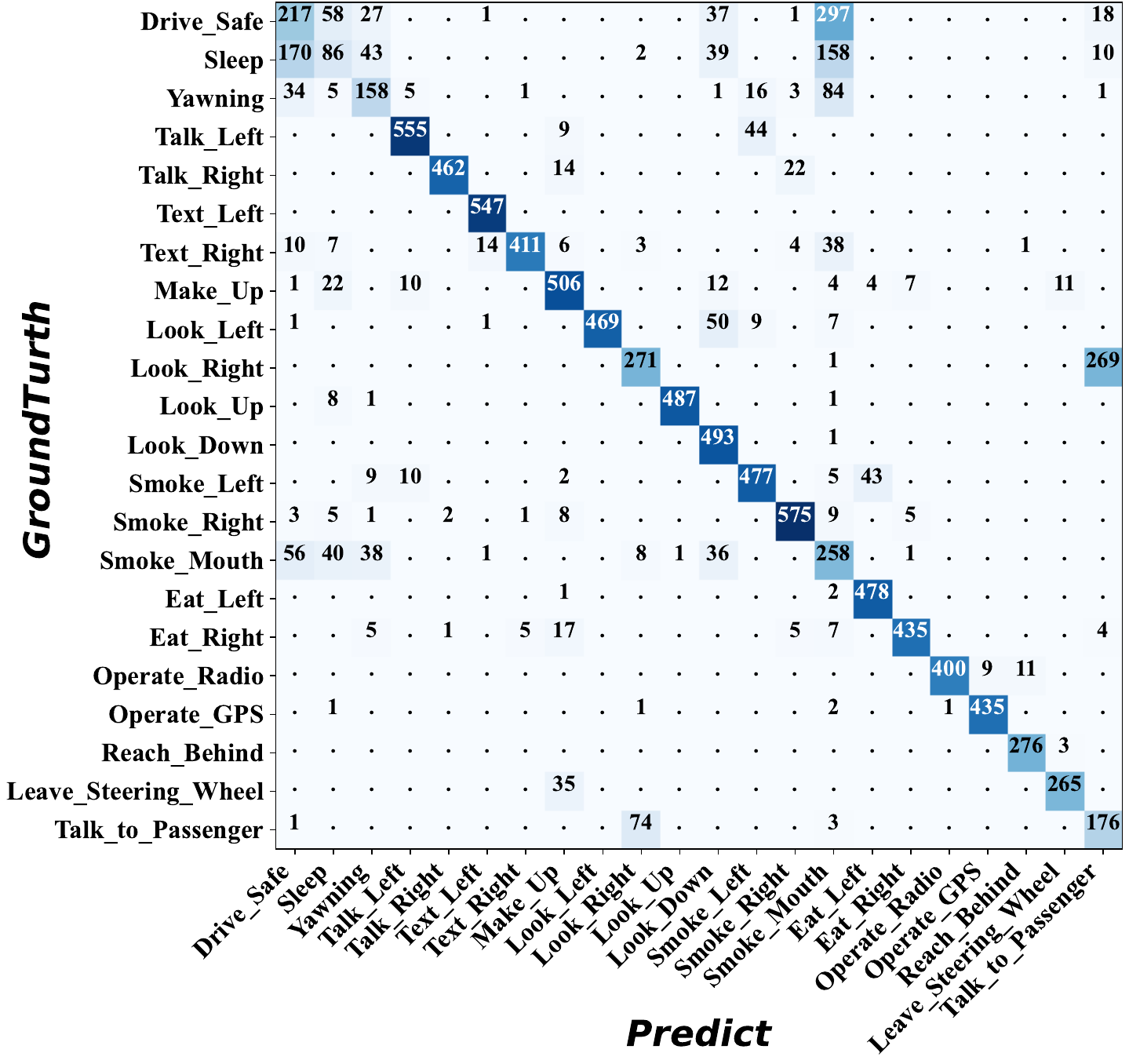}&
		\includegraphics[width=0.25\linewidth]{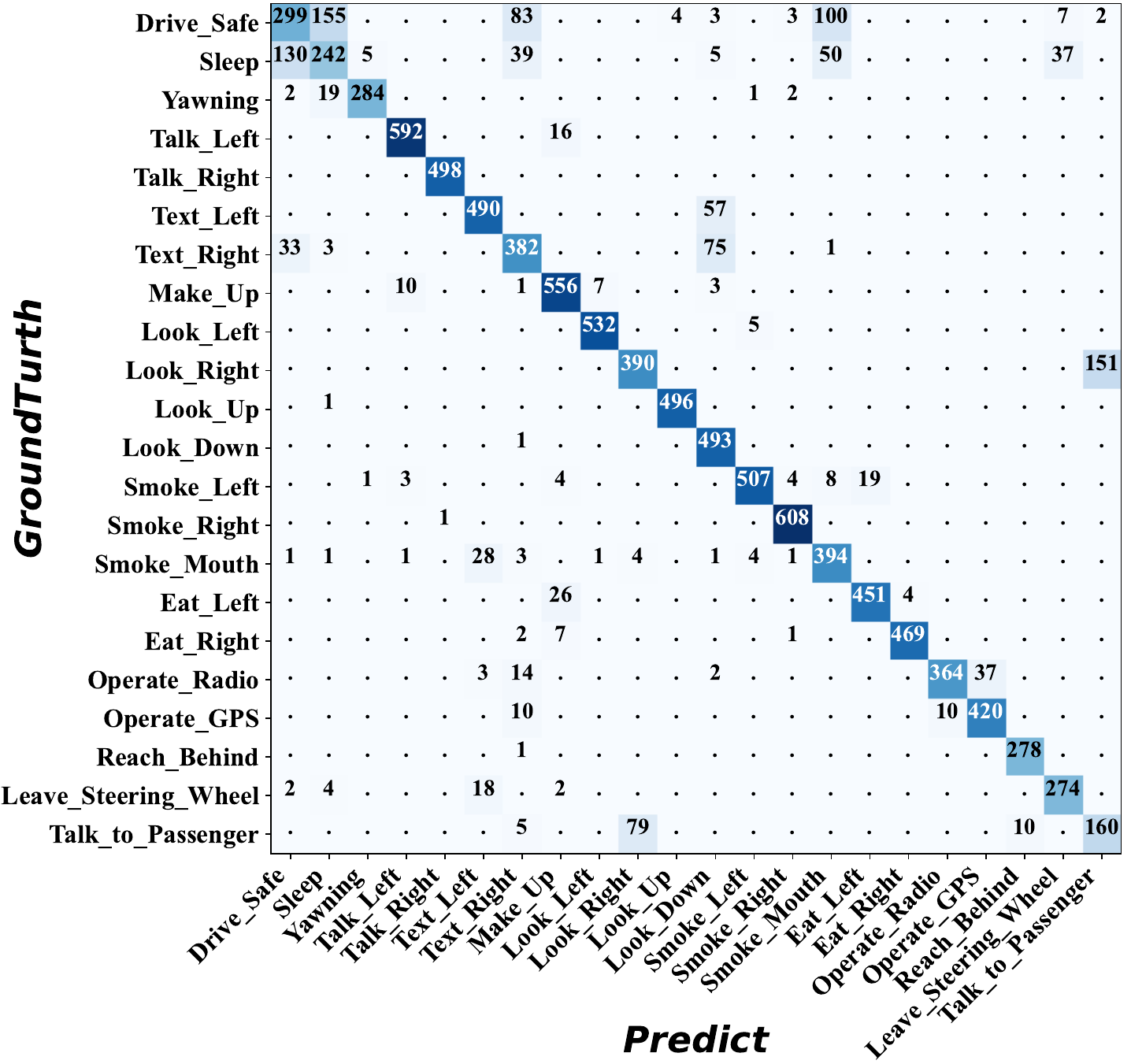}\\
		(a)&(b)&(c)&(d)
	\end{tabular}
	\caption{\modifyRS{Visualizing the class confusion matrices to explore the impact of applying GF on different categories. (a) and (b) are the cross-dataset confusion matrices based on ResNet18 on EZZ2021, with the fusion branches corresponding to $\{\mathbb{D}_{12}^*|20, \mathbb{D}_{13}^*|10, \mathbb{D}_{23}^*|5\}$ in Table \ref{Table_fusion_ezz2021}. (a) presents the results of the best single branch, $\mathbb{D}_{13}^*|10$, while (b) shows the fusion results. (c) and (d) display the multi-camera fusion results on the 100-Driver dataset based on ResNet50 in Table \ref{Table_fusion_camera}, with the fusion branches corresponding to \{$\text{D1}$,$\text{D2}$,$\text{D4}$\}. (c) shows the results of the best single branch $\text{D4}$, while (d) shows the fusion results.}}
	\label{Figure_GFConfusionMatrix}
\end{figure*}

\subsection{Multi-channel Information Gaussian Fusion}
Section \ref{MCIF} indicates the significant advantages and importance of feature fusion.
Thus, we designed a multi-channel Gaussian fusion (GF) strategy based on the S-Softmax classifier in Section \ref{Multi-channel feature fusion}.
In this section, we conduct extensive experiments to demonstrate the effectiveness of the proposed GF method.

Table \ref{Table_twoDatasetTrain_crossDatasetTest} indicates combined datasets outperform single datasets.
Hence, models trained on combined datasets are used as branches for the fusion strategy in this section.
And the all samples of EZZ2021 ($\mathbb{T}_4$) \cite{zandamelaCrossdatasetPerformanceEvaluation2022} are used for testing.
The additive fusion (AF), which is used in \cite{Wang_100Driver2023}, is just feasible when $G$ is equality for all branchs.
The results of Table \ref{Table_fusion_ezz2021},  indicate that GF is slightly superior to AF.
For different $G$, AF fails to deal with this issue, while GF can still consistently improve the accuracy of each branch fusion.
Choosing the optimal combination for fusion can maximize the advantage. 
For example, the fusion of branch combination $\{\mathbb{D}_{12}^*|20, \mathbb{D}_{13}^*|10\}$ increased the accuracy from 69.6\% to 74.4\%, while the fusion of three branches, $\{\mathbb{D}_{12}^*|20,\mathbb{D}_{13}^*|10, \mathbb{D}_{23}^*|5\}$, reached 75.1\%.

\begin{table}[!t]\scriptsize
	\setlength{\tabcolsep}{5.8pt}
	\caption{Multi-Channel Fusion Experiments On \textbf{EZZ2021}. $\mathbb{D}_{ij}^*$ Means the Combined Dataset of $\mathbb{D}_i^*$ and $\mathbb{D}_j^*$. $\mathbb{D}_{ij}^*|G$ Means the Score Level of S-Softmax is $G$. \textbf{AF} Means Additive Fusion. \textbf{GF} Means Gaussian Fusion. \ding{56} Means the Fusion Method Failure. The \textbf{Bold} Accuracy Means the Best Result Among Single Branch, \textbf{Add}, and \textbf{GF}.}
	\centering
	\begin{tabular}{l|c|c|c|c|c}
		\toprule[1.2pt]
		\parbox[c]{2.4cm}{\bf{Multi-Channel Config: \{$\mathcal{S}^1, \mathcal{S}^2,\mathcal{S}^3, \cdots$\}}}&{$\mathcal{S}^1$}&{$\mathcal{S}^2$}&{$\mathcal{S}^3$}&\textbf{AF}&	\parbox[c]{0.4cm}{\textbf{  GF} (Ours)}\\
		
		\hline 
		\multicolumn{6}{c}{\cellcolor{lightgray}\textit{\textbf{Multi-Channel Fusion of Different Dataset, with $G=5$}}}\\
		$\{\mathbb{D}_{12}^*,\mathbb{D}_{13}^*\}$ &60.2&67.7&-&\valueUP{69.8}{2.1}&\textbf{\valueUP{70.2}{2.5}}\\     
		$\{\mathbb{D}_{12}^*,\mathbb{D}_{23}^*\}$ &60.2&69.1&-&\valueUP{69.9}{0.8}&\textbf{\valueUP{70.0}{0.9}}\\    
		$\{\mathbb{D}_{13}^*,\mathbb{D}_{23}^*\}$ &67.7&69.1&-&\valueUP{70.6}{1.5}&\textbf{\valueUP{70.9}{1.8}}\\    
		\hline
		\multicolumn{6}{c}{\cellcolor{lightgray}\textit{\textbf{Multi-Channel Fusion of Different $G$, With the Same Dataset $\mathbb{D}_{12}^*$}}}\\
		$\{G\text{=}5, G\text{=}20\}$ &60.2&65.9&-&\ding{56}&\textbf{\valueUP{66.9}{1.0}}\\    
		$\{G\text{=}5, G\text{=}10\}$ &67.7&69.6&-&\ding{56}&\textbf{\valueUP{70.0}{0.4}}\\   
		$\{G\text{=}5, G\text{=}20\}$ &69.1&65.2&-&\ding{56}&\textbf{\valueUP{69.6}{0.5}}\\   
		\hline
		\multicolumn{6}{c}{\cellcolor{lightgray}\textit{\textbf{Multi-Channel Fusion of Different Dataset and Different  $G$}}}\\
		$\{\mathbb{D}_{12}^*|20, \mathbb{D}_{23}^*|5\}$ &65.9&69.1&-&\ding{56}&\textbf{\valueUP{72.8}{3.7}}\\  
		$\{\mathbb{D}_{12}^*|20, \mathbb{D}_{13}^*|10\}$ &65.9&69.6&-&\ding{56}&\textbf{\valueUP{74.4}{4.8}}\\   
		$\{\mathbb{D}_{13}^*|10, \mathbb{D}_{23}^*|5\}$ &69.6&69.1&-&\ding{56}&\textbf{\valueUP{72.8}{3.2}}\\    
		$\{\mathbb{D}_{12}^*|20,\mathbb{D}_{13}^*|10, \mathbb{D}_{23}^*|5\}$
		&65.9&69.6&69.1&\ding{56}&\valueUP{\textbf{75.1}}{5.5}\\ 
		\bottomrule[1.2pt]
	\end{tabular}
	\label{Table_fusion_ezz2021}%
\end{table}%
\begin{table}[bt]\scriptsize
	\setlength{\tabcolsep}{2pt}
	\caption{Accuracy of Multi-Backbone Gaussian Fusion (\textbf{GF}) Based on Cross-Camera and Cross-Vehicle Config of \textbf{100-Driver}. The R, E, S and G Means the Backbone of ResNet50, EfficientNetB0, ShuffleNetV2 and GhostNetV1, respectively. The \textbf{Bold} Accuracy Means the Best Result Among $\mathcal{S}_1$, $\mathcal{S}_2$ and \textbf{GF}. The $\text{M}\rightarrow\text{M}$ Means the CNN Trained on the Train Subset of $\text{Mazda}$ and Tested on the Test Subset of $\text{Mazda}$.}
	\centering
	\begin{tabular}{l|ccc|ccc|ccc|ccc}
		\bottomrule[1.2pt]
		\multirow{2}{1.7cm}{{\bf{Fusion Branch: \{$\mathcal{S}_1,\mathcal{S}_2$\}}}}
		&\multicolumn{3}{c}{\cellcolor{lightgray}D1}&\multicolumn{3}{c}{\cellcolor{lightgray}D2}&\multicolumn{3}{c}{\cellcolor{lightgray}D3}&\multicolumn{3}{c}{\cellcolor{lightgray}D4}\\
		&$\mathcal{S}_1$&$\mathcal{S}_2$&\textbf{GF}&$\mathcal{S}_1$&$\mathcal{S}_2$&\textbf{GF}&$\mathcal{S}_1$&$\mathcal{S}_2$&\textbf{GF}&$\mathcal{S}_1$&$\mathcal{S}_2$&\textbf{GF}\\
		\hline
		D1: \{R, E\}&76.5&73.7&\textbf{\valueUP{77.1}{0.6}}&56.6&54.4&\textbf{\valueUP{58.1}{1.5}}&27.8&23.6&\textbf{\valueUP{28.4}{0.6}}&4.8&\textbf{6.5}&\valueDown{4.5}{0.3}\\     
		
		D2: \{R, E\}&\textbf{47.0}&30.9&\valueDown{39.9}{7.1}&73.4&\textbf{78.9}&\valueDown{78.6}{0.3}&47.0&44.8&\textbf{\valueUP{50.5}{3.5}}&\textbf{8.2}&5.9&\valueDown{6.9}{1.3}\\     
		
		D3: \{R, E\}&\textbf{26.1}&20.7&\valueDown{24.0}{2.1}&\textbf{46.4}&41.3&\valueDown{45.6}{0.8}&77.3&80.2&\textbf{\valueUP{80.6}{0.4}}&13.1&12.4&\textbf{\valueUP{14.4}{1.3}}\\     
		
		D4: \{R, G\}&3.5&\textbf{4.7}&\valueDown{3.1}{1.6}&\textbf{4.8}&4.5&\valueDown{4.2}{0.9}&\textbf{17.3}&9.5&\valueDown{16.1}{1.2}&80.5&75.5&\textbf{\valueUP{81.4}{0.9}}\\     
		\bottomrule
		
		\multirow{2}{1.7cm}{{\bf{Fusion Branch: \{$\mathcal{S}_1,\mathcal{S}_2$\}}}}&\multicolumn{3}{c}{\cellcolor{lightgray}M$\rightarrow$M}&\multicolumn{3}{c}{\cellcolor{lightgray}M$\rightarrow$H}&\multicolumn{3}{c}{\cellcolor{lightgray}M$\rightarrow$A}&\multicolumn{3}{c}{\cellcolor{lightgray}M$\rightarrow$L}\\
		&$\mathcal{S}_1$&$\mathcal{S}_2$&\textbf{GF}&$\mathcal{S}_1$&$\mathcal{S}_2$&\textbf{GF}&$\mathcal{S}_1$&$\mathcal{S}_2$&\textbf{GF}&$\mathcal{S}_1$&$\mathcal{S}_2$&\textbf{GF}\\
		\hline
		D1: \{R, E\}&72.6&74.4&\textbf{\valueUP{76.8}{2.4}}&61.4&61.6&\textbf{\valueUP{66.1}{4.5}}&\textbf{45.8}&26.2&\valueDown{41.6}{4.2}&58.8&55.7&\textbf{\valueUP{61.1}{2.3}}\\     
		
		D2: \{R, E\}&75.0&75.0&\textbf{\valueUP{75.7}{0.7}}&50.3&48.7&\textbf{\valueUP{54.3}{4.0}}&56.4&50.4&\textbf{\valueUP{60.0}{3.6}}&66.0&61.4&\textbf{\valueUP{66.5}{0.5}}\\     
		
		D3: \{R, E\}&70.8&76.1&\textbf{\valueUP{77.0}{0.9}}&61.6&56.3&\textbf{\valueUP{62.4}{0.8}}&\textbf{34.5}&14.7&\valueDown{33.3}{1.2}&58.0&45.3&\textbf{\valueUP{58.3}{0.3}}\\     
		
		D4: \{S,   E\}&63.4&73.6&\textbf{\valueUP{74.4}{0.8}}&57.5&61.5&\textbf{\valueUP{63.4}{0.7}}&\textbf{56.4}&19.9&\valueDown{42.3}{14.1}&52.4&59.4&\textbf{\valueUP{59.8}{0.4}}\\   
		\bottomrule
		
		\multirow{2}{1.7cm}{{\bf{Fusion Branch: \{$\mathcal{S}_1,\mathcal{S}_2$\}}}}&\multicolumn{3}{c}{\cellcolor{lightgray}Se$\rightarrow$Se}&\multicolumn{3}{c}{\cellcolor{lightgray}Se$\rightarrow$SUV}&\multicolumn{3}{c}{\cellcolor{lightgray}Se$\rightarrow$Van}&&&\\
		&$\mathcal{S}_1$&$\mathcal{S}_2$&\textbf{GF}&$\mathcal{S}_1$&$\mathcal{S}_2$&\textbf{GF}&$\mathcal{S}_1$&$\mathcal{S}_2$&\textbf{GF}&&&\\
		\cline{1-10}
		D1: \{R, E\}&70.3&75.0&\textbf{\valueUP{75.3}{0.3}}&65.6&68.2&\textbf{\valueUP{71.4}{3.2}}&\textbf{40.7}&19.8&\valueDown{36.8}{3.9}&&&\\   
		D2: \{R, E\}&77.1&73.1&\textbf{\valueUP{77.6}{0.5}}&49.5&\textbf{51.4}&\valueDown{50.7}{0.7}&\textbf{55.7}&33.0&\valueDown{55.3}{0.4}&&&\\    
		
		D3: \{R, E\}&73.0&75.4&\textbf{\valueUP{75.6}{0.2}}&64.9&64.3&\textbf{\valueUP{67.1}{2.2}}&\textbf{53.0}&15.5&\valueDown{51.6}{1.4}&&&\\    
		
		D4: \{R, E\}&73.6&72.6&\textbf{\valueUP{76.3}{2.7}}&72.3&67.8&\textbf{\valueUP{73.0}{0.7}}&\textbf{73.7}&38.3&\valueDown{71.7}{2.0}&&&\\   
		
		\toprule[1.2pt]
	\end{tabular}
	\label{Table_fusion_backbone}%
\end{table}%
Table \ref{Table_fusion_backbone} presents the multi-backbone fusion results of cross-camera and cross-vehicle based on the 100-Driver dataset.
The results indicate that when each branch is in a good state, the fusion strategy can significantly improve accuracy.
For instance, whether the non-cross-vehicle testing of $\text{M}\rightarrow \text{M}$, $\text{Se}\rightarrow \text{Se}$, or the cross-vehicle settings of $\text{M}\rightarrow \text{H}$, $\text{M}\rightarrow \text{L}$, and $\text{Se}\rightarrow \text{SUV}$, all accuracy showed significant improvements because of GF. 
Similarly, in $\text{D1}\rightarrow \text{D1}$, $\text{D1}\rightarrow \text{D2}$, and $\text{D1}\rightarrow \text{D3}$ of D1: \{R, E\}, the same trend was observed.
When one of the branches is performing poorly, GF fails guarantee improved accuracy. For example, in the results of $\text{M}\rightarrow \text{A}$, the accuracy of D1: \{R, E\}, D3: \{R, E\}, and D4: \{S, E\} with GF is slightly lower than that of $\mathcal{S}_1$. 
However, GF tends towards the better-performing branch, and the poor branches have less impact on the well-performing branches.
Similarly, this is also evident in $\text{Se}\rightarrow \text{Van}$.
This is a characteristic of GF, similar to what is observed in the Kalman Filter \cite{faragherUnderstandingBasisKalman2012}.

\begin{table}[!t]\scriptsize
	\setlength{\tabcolsep}{4.2pt}
	\caption{Comparing the \textbf{Add} Fusion \cite{Wang_100Driver2023} with Gaussian Fusion (\textbf{GF}) of Multi-Camrea Branch Inputs on \textbf{100-Driver}. The \textbf{Bold} Accuracy Means the Best Result is Ours \textbf{GF} Instead of Wang's Method \cite{Wang_100Driver2023}.}
	\centering
	\begin{tabular}{ll|cccccc}
		\toprule[1.2pt]
		\parbox[c]{1.0cm}{\bf{MULTI-CAMERA FUSION }}&\parbox[l]{0.1cm}&\parbox[l]{0.7cm}{Res\\Net50}&\parbox[c]{0.7cm}{Mobile\\NetV3}&\parbox[c]{0.7cm}{Shuffle\\NetV2}&\parbox[c]{0.7cm}{Squee-\\zeNet}&\parbox[c]{0.7cm}{Efficient\\NetB0}&\parbox[c]{0.7cm}{Ghost\\NetV1}\\
		\hline
		\multicolumn{2}{c|}{\#Params (M)}&23.5&4.2&1.3&0.7&4.0&3.9\\
		\hline
		\multicolumn{2}{c|}{MACs (MB)}&4109.5&224.2&147.8&737.4&400.4&146.9\\
		\multicolumn{8}{c}{\cellcolor{lightgray}\textit{\textbf{Fusion of two cameras}}}\\
		\multirow{2}{*}{\{$\text{D1}$,$\text{D2}$\}}& \cite{Wang_100Driver2023}&73.8&77.1&73.2&82.1&81.7&74.8\\
		&\textbf{GF}&\valueUP{\textbf{80.1}}{3.6}&\valueUP{75.9}{2.4}&\valueUP{\textbf{75.1}}{5.5}&\valueUP{74.7}{3.4}&\valueUP{80.9}{2.0}&\valueUP{\textbf{77.7}}{4.9}\\ \hline
		\multirow{2}{*}{\{$\text{D1}$,$\text{D3}$\}} &\cite{Wang_100Driver2023}&74.7&81.9&75.1&80.4&83.6&78.0\\ &\textbf{GF}&\valueUP{\textbf{80.3}}{3.0}&\valueUP{\textbf{83.1}}{4.1}&\valueUP{\textbf{78.7}}{5.6}&\valueUP{80.0}{4.6}&\valueUP{82.8}{2.6}&\valueUP{\textbf{82.8}}{7.5}\\ \hline
		\multirow{2}{*}{\{$\text{D1}$,$\text{D4}$\}}&\cite{Wang_100Driver2023}&82.5&83.8&79.9&86.2&87.9&83.7\\
	   &\textbf{GF}&\valueUP{\textbf{86.4}}{5.9}&\valueUP{\textbf{86.1}}{6.8}&\valueUP{\textbf{85.1}}{11.0}&\valueUP{83.9}{11.4}&\valueUP{85.6}{5.4}&\valueUP{83.6}{8.1}\\ \hline
		\multirow{2}{*}{\{$\text{D2}$,$\text{D3}$\}}&\cite{Wang_100Driver2023}&72.8&77.9&72.2&83.6&80.0&75.8\\
		 &\textbf{GF}&\valueUP{\textbf{82.3}}{5.0}&\valueUP{\textbf{81.9}}{2.9}&\valueUP{\textbf{77.7}}{4.6}&\valueUP{76.7}{1.3}&\valueUP{\textbf{85.4}}{5.2}&\valueUP{\textbf{81.2}}{5.9}\\ \hline
		\multirow{2}{*}{\{$\text{D2}$,$\text{D4}$\}}&\cite{Wang_100Driver2023}&78.7&82.6&76.1&85.2&83.2&80.1\\
		 &\textbf{GF}&\valueUP{\textbf{86.1}}{5.6}&\valueUP{\textbf{84.5}}{5.2}&\valueUP{\textbf{83.8}}{9.7}&\valueUP{82.4}{9.9}&\valueUP{\textbf{87.9}}{7.7}&\valueUP{\textbf{84.5}}{9.0}\\ \hline
		\multirow{2}{*}{\{$\text{D3}$,$\text{D4}$\}}&\cite{Wang_100Driver2023}&80.8&82.2&76.7&86.2&82.9&82.4\\
		 &\textbf{GF}&\valueUP{\textbf{86.0}}{4.5}&\valueUP{\textbf{85.3}}{6.0}&\valueUP{\textbf{83.1}}{9.0}&\valueUP{83.7}{8.3}&\valueUP{\textbf{85.6}}{5.4}&\valueUP{\textbf{82.8}}{7.3}\\
		\hline
		\multicolumn{8}{c}{\cellcolor{lightgray}\textit{\textbf{Fusion of three cameras}}}\\
		\multirow{2}{*}{\{$\text{D1}$,$\text{D2}$,$\text{D3}$\}}&\cite{Wang_100Driver2023}&76.8&83.5&77.2&82.9&84.8&82.1\\
		 &\textbf{GF}&\valueUP{\textbf{82.7}}{0.4}&\valueUP{\textbf{83.6}}{0.5}&\valueUP{\textbf{78.7}}{0.0}&\valueUP{80.1}{0.1}&\valueUP{\textbf{86.0}}{0.6}&\valueUP{\textbf{83.7}}{0.9}\\ \hline
		
		\multirow{2}{*}{\{$\text{D1}$,$\text{D2}$,$\text{D4}$\}}&\cite{Wang_100Driver2023}&82.5&86.1&78.4&84.9&87.8&85.1\\
		 &\textbf{GF}&\valueUP{\textbf{87.3}}{0.9}&\valueUP{86.1}{0.1}&\valueDown{\textbf{85.0}}{0.1}&\valueUP{84.1}{0.2}&\valueUP{\textbf{89.0}}{1.1}&\valueUP{\textbf{85.7}}{1.2}\\  \hline
		
		\multirow{2}{*}{\{$\text{D1}$,$\text{D3}$,$\text{D4}$\}}&\cite{Wang_100Driver2023}&83.3&84.8&79.7&86.5&86.8&85.0\\
		 &\textbf{GF}&\valueUP{\textbf{86.7}}{0.3}&\valueUP{\textbf{87.6}}{1.5}&\valueUP{\textbf{85.8}}{0.7}&\valueUP{86.1}{2.2}&\valueUP{\textbf{87.5}}{1.9}&\valueUP{\textbf{86.6}}{3.0}\\    \hline 
         \multirow{2}{*}{\{$\text{D2}$,$\text{D3}$,$\text{D4}$\}}&\cite{Wang_100Driver2023}&83.0&84.6&77.8&84.7&84.5&83.6\\
	    &\textbf{GF}&\valueUP{\textbf{87.3}}{1.2}&\valueUP{\textbf{88.4}}{3.1}&\valueUP{\textbf{84.9}}{1.1}&\valueUP{84.7}{1.0}&\valueUP{\textbf{89.1}}{1.2}&\valueUP{\textbf{86.2}}{1.7}\\   
		\hline 
		\multicolumn{8}{c}{\cellcolor{lightgray}\textit{\textbf{Fusion of four cameras }}}\\
		\multirow{2}{*}{\{$\text{D1}$,$\text{D2}$,$\text{D3}$,$\text{D4}$\}}&{\cite{Wang_100Driver2023}}&84.4&86.9&80.0&83.5&86.9&86.2\\
		&\textbf{GF}&\valueDown{\textbf{87.0}}{0.3}&\valueUP{\textbf{89.0}}{0.6}&\valueUP{\textbf{86.0}}{0.2}&\valueDown{\textbf{85.8}}{0.3}&\valueUP{\textbf{89.4}}{0.3}&\valueUP{\textbf{88.8}}{2.2}\\     
		
		\bottomrule[1.2pt]
	\end{tabular}
	\label{Table_fusion_camera}%
\end{table}%
Table \ref{Table_fusion_camera} compares the multi-camera fusion results of AF based on Softmax \cite{Wang_100Driver2023} with the GF based on S-Softmax.
For ResNet50, MobileNetV3, ShuffleNetV2, and GhostNetV1, the accuracy of GF significantly outperforms the results reported by Wang \emph{et al.} \cite{Wang_100Driver2023}.
SqueezeNet achieves the best result when four camera branches are fused.
Deep models tend to overfit more easily on datasets with limited diversity \cite{pasupaComparisonShallowDeep2016}, while S-Softmax helps tackle this issue.
Furthermore, the fusion of more cameras results in more stable improvements because the GF tends towards the better-performing branches.
Therefore, GF allows for more effective utilization of useful information from multiple branches and serves as a reference for multimodal fusion and global-local fusion.
\modifyRS{The confusion matrices pre- and post-fusion indicate that GF can improve the accuracy of the vast majority of driver behavior categories. In Fig. \ref{Figure_GFConfusionMatrix} (a) and (b), except for "Drive Safety," which is more prone to being misclassified as "Text Left," all categories have shown improvement. "Text Right" and "Talk Right" are more accurately classified, while "Text Left" is misclassified as "Talk Left" rather than "Hair \& Makeup." This is because holding a phone in the left hand is more likely to be obstructed by the body, but evidently, the relevant features are more easily recognized. In Fig. \ref{Figure_GFConfusionMatrix} (c) and (d), recognition rates across the majority of categories in the 100-Driver dataset have improved.}

\section{CONCLUSION}\label{section:6}
\modifyRS{The limited diversity of the dataset, along with the use of the Softmax classifier and One-Hot label, increases the susceptibility of CNNs to noise traps. This paper proposes the S-Softmax classifier and DGSS, which simultaneously ahieves label smoothing and label relaxation, aiming to mitigate the overconfidence of models induced by Softmax and One-Hot labels.
Experiments demonstrate that S-Softmax@DGSS improves ResNet18's performance to 82.19\%, 63.39\%, and 74.04\% on SFDDD, AUCDD, and 100-DriverM, respectively, outperforming existing label smoothing methods. Similar enhancements are observed across other models.
Furthermore, the GF method achieves state-of-the-art results of 75.1\% and 89.4\% on EZZ2021 and 100-Driver, respectively, which is of significant importance for distracted driving detection in NDS. While S-Softmax@DGSS effectively reduces model overconfidence by mitigating the impact of noise, it does not enhance the model's feature capture capability to adequately address the challenge posed by significant differences in camera perspectives. Achieving reliable DMS in NDS remains an ongoing endeavor. Future efforts will explore combining S-Softmax with CLIP to bolster CNNs' capacity to capture pertinent driver behavior features in NDS.}

\scriptsize
\bibliography{ref}
\bibliographystyle{IEEEtran}
\vspace{-7 mm} 
\begin{IEEEbiography}[{\includegraphics[width=1in,height=1.25in,clip,keepaspectratio]{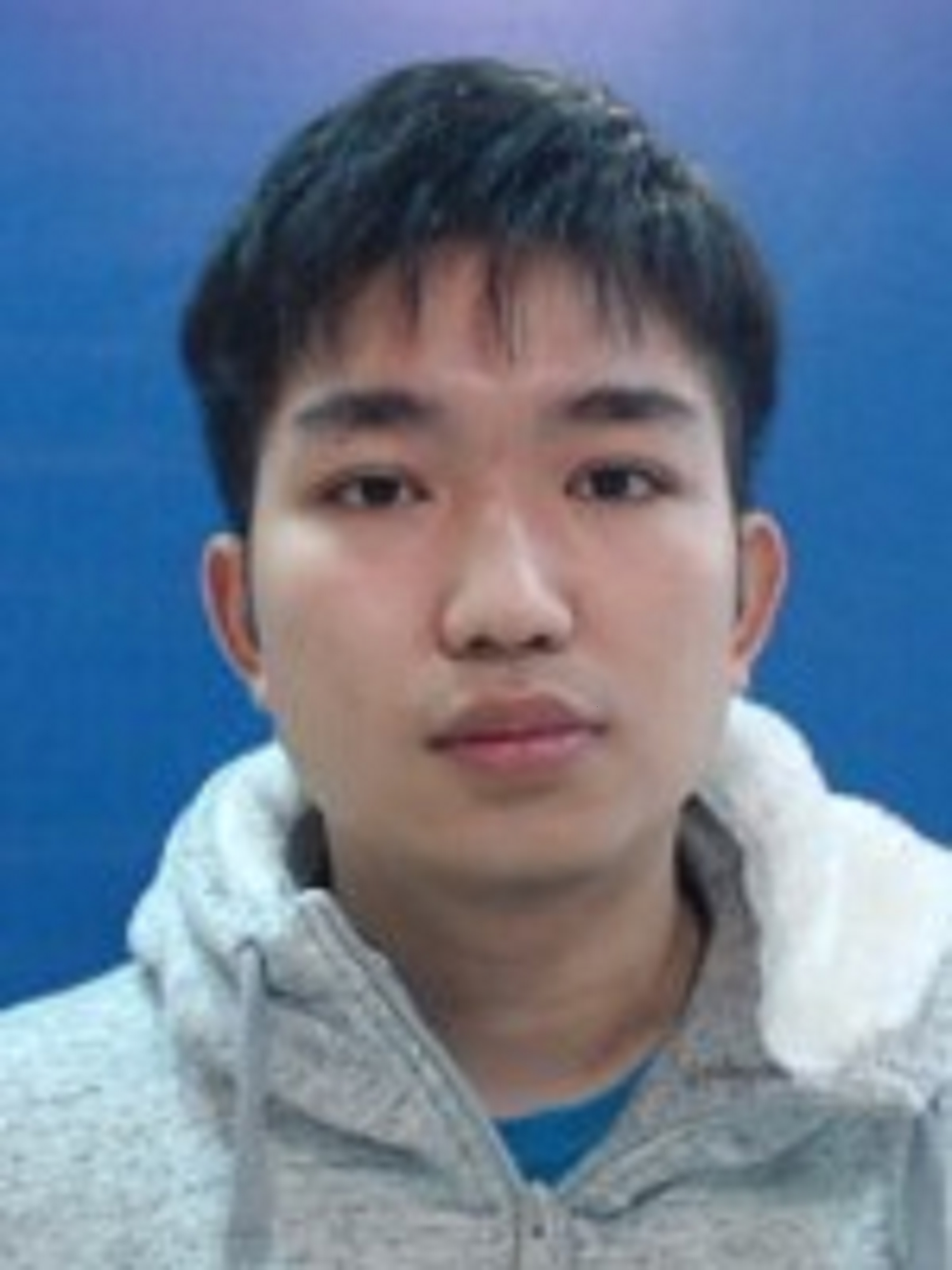}}]{Cong Duan} received the Bachelor’ Degree from Dalian University of Technology, Dalian, China, in 2017 and began to study for a master’s degree in Hunan University. Since 2021 he has been working toward a Ph.D. degree in the State Key Laboratory of Advanced Design and Manufacturing for Vehicle Body, College of Mechanical and Vehicle Engineering, Hunan University, Changsha, China. His research interests is vehicle intelligent driving technology, including imaging processing, object visual tracking and advanced driving assistance technology.
\end{IEEEbiography}
\vspace{-9 mm} 
\begin{IEEEbiography}[{\includegraphics[width=1in,height=1.25in,clip,keepaspectratio]{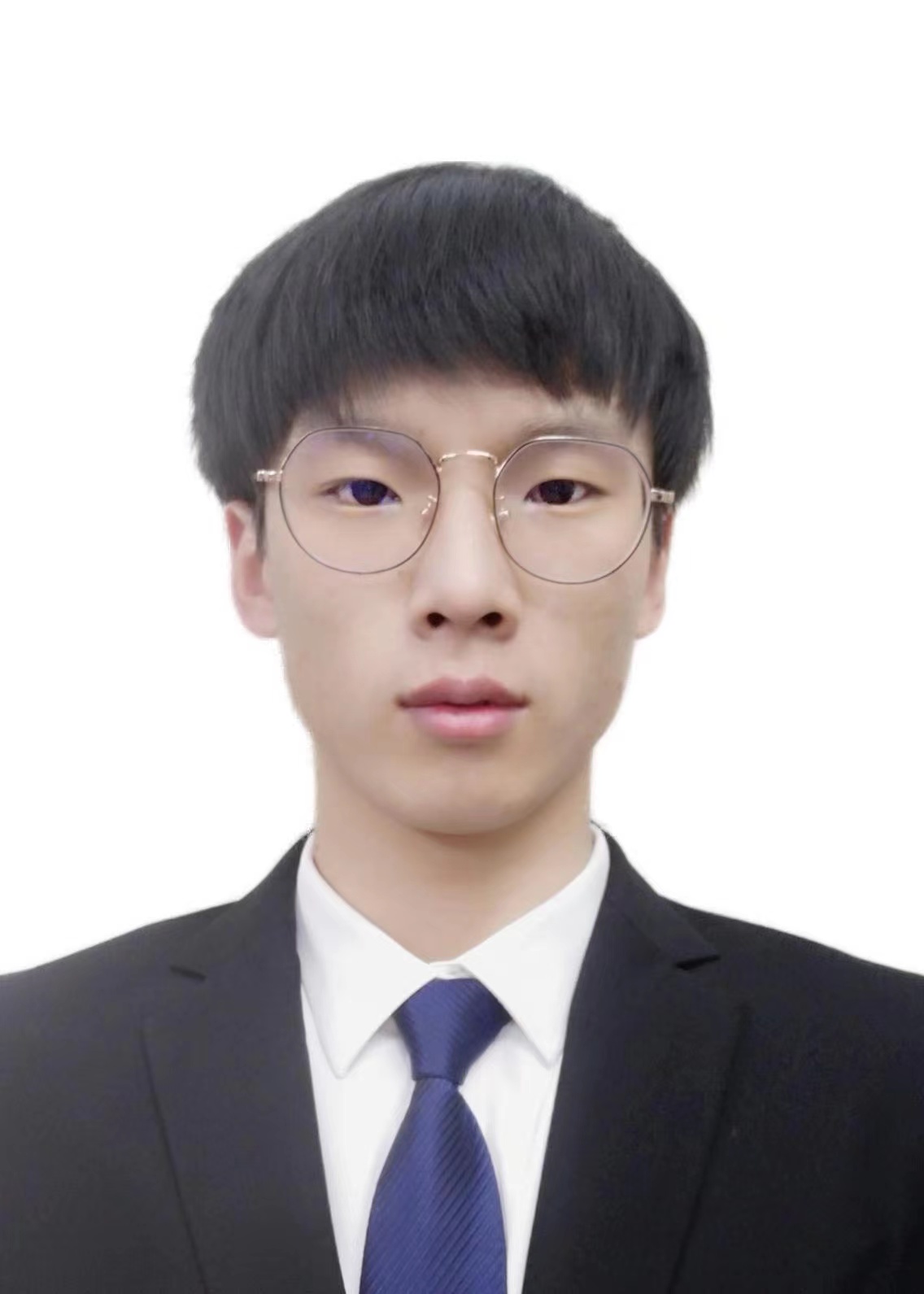}}]{Zixuan Liu} obtained a Bachelor's degree from Northeast Forestry University in 2022. Since 2022, he has studied for a master's degree in the State Key Laboratory for Advanced Design and Manufacturing of Automobile Body, College of Mechanical and Vehicle Engineering of Hunan University. His research interests include computer vision, deep learning, and advanced driving assistance technologies.
\end{IEEEbiography}
\vspace{-7 mm} 
\begin{IEEEbiography}[{\includegraphics[width=1in,height=1.25in,clip,keepaspectratio]{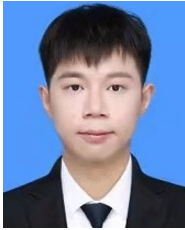}}]{Jiahao Xia} (Graduate Student Member, IEEE) received the B.Eng.degree from the Wuhan University of Technology, Wuhan, China, in 2017, and the M.Eng.degree from Hunan University, Changsha, China, in 2020. He is currently pursuing the Ph.D degree with the School of Electrical and Data Engineering, University of Technology Sydney, Ultimo, NSW, Australia. 
His current research interests include vision transformer, unsupervised learning,and graph neural networks.
\end{IEEEbiography}
\vspace{-7 mm} 
\begin{IEEEbiography}[{\includegraphics[width=1in,height=1.25in,clip,keepaspectratio]{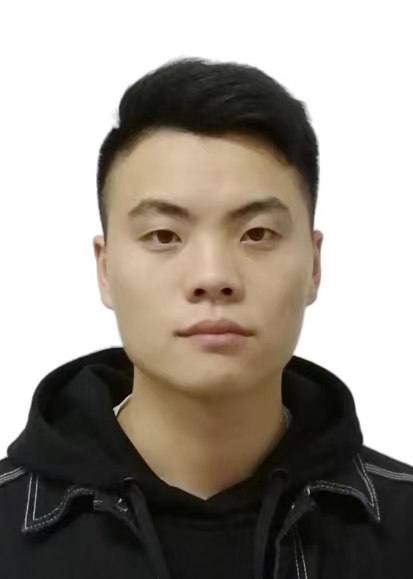}}]{Minghai Zhang}  received the B.E degree in engineering mechanics from Northeastern University, Shenyang, China, in 2019. He is working toward the Ph.D. degree in the State Key Laboratory of Advanced Design and Manufacturing for Vehicle Body, College of Mechanical and Vehicle Engineering, Hunan University. His research interests include motion planning, control and automatic driving.
\end{IEEEbiography}
\vspace{-7 mm} 
\begin{IEEEbiography}[{\includegraphics[width=1in,height=1.25in,clip,keepaspectratio]{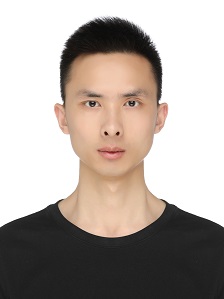}}]{Jiacai Liao} received the M.E degree in vehicle engineering from Hunan University of Mechanical and Vehicle Engineering, Changsha, China, in 2017. He got his Ph.D. degree in the State Key Laboratory of Advanced Design and Manufacturing for Vehicle Body, College of Mechanical and Vehicle Engineering, Hunan University, Changsha, China, in 2022. He has been awarded a scholarship under the State Scholarship Fund to pursue study for 14 months at Concordia University, Canada, as a Visiting Ph.D. Student between March 2021 and May 2022. At present, his a full-time teacher and researcher in the College of Automotive and Mechanical Engineering, Changsha University of Science \& Technology, Changsha, China. His research interests include semantic segmentation, self-driving cars, and intelligent transportation systems.
\end{IEEEbiography}
\vspace{-7 mm} 
\begin{IEEEbiography}[{\includegraphics[width=1in,height=1.25in,clip,keepaspectratio]{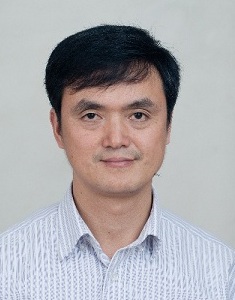}}]{Libo Cao} received the B.S. degrees in vehicle engineering from the University of Hunan, Changsha, in 1989 and the Ph.D. degree in mechanical engineering from Hunan University, Changsha, China, in 2002.
	From 2002, he was a doctoral supervisor with the State Key Laboratory of Advanced Design and Manufacturing for Vehicle Body. He has been a visiting scholar between 2003 and 2004 at University of Technology Berlin, Germany. He was joint study at Wayne State University, USA. He is the author of one book, more than 100 articles, and more than 5 Chinese national inventions. His research interests include Active safety and advance driver assisted system, injury biomechanics, passive safety, and self-driving. He is a reviewer of the Journal of automotive safety and engery, and Automotive technology.
\end{IEEEbiography}
\end{document}